\documentclass[conference]{IEEEtran}
\IEEEoverridecommandlockouts
\usepackage{cite}
\usepackage{amsmath,amssymb,amsfonts}
\usepackage{algorithmic}
\usepackage{graphicx}
\usepackage{textcomp}
\usepackage{xcolor}

\usepackage{authblk} 
\newcommand*{\affaddr}[1]{#1} 

\newcommand*{\email}[1]{\texttt{#1}}

\usepackage{fancyhdr}
\usepackage[hyphens]{url}
\usepackage{comment}
\usepackage{soul}
\usepackage{tikz}
\usepackage{multirow}
\usepackage{listings}
\usepackage{cite}
\usepackage{amsmath,amssymb,amsfonts}
\usepackage{algorithmic}
\usepackage{graphicx}
\usepackage{textcomp}
\usepackage{xcolor}
\usepackage{fancyhdr}
\usepackage[hyphens]{url}

\usepackage{soul}
\usepackage{hhline}
\usepackage{cite}

\usepackage{mathptmx} 
\usepackage{xspace}
\usepackage{booktabs}
\usepackage{threeparttable}
\usepackage{multirow}
\usepackage{multicol}
\usepackage{array}
\usepackage{color}
\usepackage{pifont}
\usepackage{lipsum}
\newcommand\mynumb[1]{\ifcase#1 \or \ding{182}\or \ding{183}\or
 \ding{184}\or \ding{185}\or \ding{186}\or \ding{187}%
 \or \ding{188}\or \ding{189}\or \ding{190}\or \ding{191}\else *\fi\relax}

\usepackage{CJK}
\usepackage{amsmath}
\usepackage{makecell}
\usepackage{amsmath}  
\usepackage{graphicx} 
\usepackage{cite}
\newcommand{\systemname}{\mbox{\text{TIMELY}}\xspace}

\def\BibTeX{{\rm B\kern-.05em{\sc i\kern-.025em b}\kern-.08em
    T\kern-.1667em\lower.7ex\hbox{E}\kern-.125emX}}
\begin{document}

\title{TIMELY: Pushing Data Movements and Interfaces in PIM Accelerators Towards Local and in Time Domain
\thanks{This work was supported in part by NIH R01HL144683 and NSF 1838873, 1816833, 1719160, 1725447, 1730309. }
}

\author{%
Weitao Li$^{1,3}$, Pengfei Xu$^{1}$, Yang Zhao$^{1}$, Haitong Li$^{2}$, Yuan Xie$^{3}$, Yingyan Lin$^{1}$\\
\affaddr{$^{1}$Department of Electrical and Computer Engineering, Rice University, Houston, TX, USA}\\
\affaddr{$^{2}$Department of Electrical Engineering,
Stanford University, Stanford, CA, USA}\\
\affaddr{$^{3}$ Department of Electrical and Computer Engineering, University of California, Santa Barbara, CA, USA}\\
\email{$^{1}$\{weitaoli,px5,zy34,yingyan.lin\}@rice.edu}\\
\email{$^{2}$\{haitongl\}@stanford.edu~~~~$^{3}$\{weitaoli,yuanxie\}@ucsb.edu}\\
}

\maketitle

\begin{abstract}
\sloppy {Resistive-random-access-memory (ReRAM) based processing-in-memory (R$^2$PIM) accelerators show promise in bridging the gap between Internet of Thing devices' constrained resources and
\textcolor[rgb]{0,0,0}{Convolutional/}Deep Neural Networks' (\textcolor[rgb]{0,0,0}{CNNs/}DNNs') prohibitive energy cost.  Specifically,  R$^2$PIM accelerators enhance energy efficiency by eliminating the cost of weight movements and improving the computational density through ReRAM's high density. However, the energy efficiency is still limited by the dominant energy cost of input and partial sum (Psum) movements and the cost of digital-to-analog (D/A) and analog-to-digital (A/D) interfaces. In this work, we identify three energy-saving opportunities in R$^2$PIM accelerators: analog data locality, time-domain interfacing, and input access reduction, 
and propose an innovative R$^2$PIM accelerator called \systemname, with three key contributions:
(1) TIMELY adopts analog local buffers (ALBs) within ReRAM crossbars to greatly enhance the data locality, minimizing the energy overheads of both input and Psum movements; (2) TIMELY largely reduces the energy of each single D/A (and A/D) conversion and the total number of conversions by using 
time-domain interfaces (TDIs) and the employed ALBs, respectively; (3) we develop an only-once input read (O$^2$IR) mapping method to further decrease the energy of input accesses and the number of D/A conversions.    
The evaluation with more than 10 CNN/DNN models and various chip configurations 
shows that, \systemname outperforms the baseline R$^2$PIM accelerator\textcolor[rgb]{0,0,0}{, PRIME,} by one order of magnitude in energy efficiency while maintaining better computational density (up to 31.2$\times$) and throughput (up to 736.6$\times$). Furthermore, comprehensive studies are performed to evaluate the  effectiveness of the proposed ALB, TDI, and O$^2$IR innovations in terms of energy savings and area reduction. }
\end{abstract}











\section{Introduction}
While deep learning-powered Internet of Things (IoT) devices promise to revolutionize the way we live and work by enhancing our ability to recognize, analyze, and classify the world around us, this revolution has yet to be unleashed.
IoT devices \textcolor[rgb]{0,0,0}{-- such as smart phones, smart sensors, and drones --} have limited energy and computation resources since they are battery-powered and have a small form factor. On the other hand, high-performance Convolutional/Deep Neural Networks (CNNs/DNNs) come at a cost of prohibitive energy consumption \cite{wang2019dual} and can have hundreds of layers~\cite{wang2019e2} and tens of millions of parameters~\cite{wu2018deep,liu2018adadeep}. Therefore, CNN/DNN-based applications can drain the battery of an IoT device very quickly if executed frequently \cite{yang2017designing}, and requires an increase in form factor for storing and executing  CNNs/DNNs~\cite{7011421,ISAAC}. 
The situation continues to worsen due to the fact that CNNs/DNNs are becoming increasingly complex as they are designed to solve more diverse and bigger tasks \cite{Google-multimodel}.
\begin{figure*}[!t]
	\centering
	\vspace{-15pt}
	\includegraphics[width=172mm]{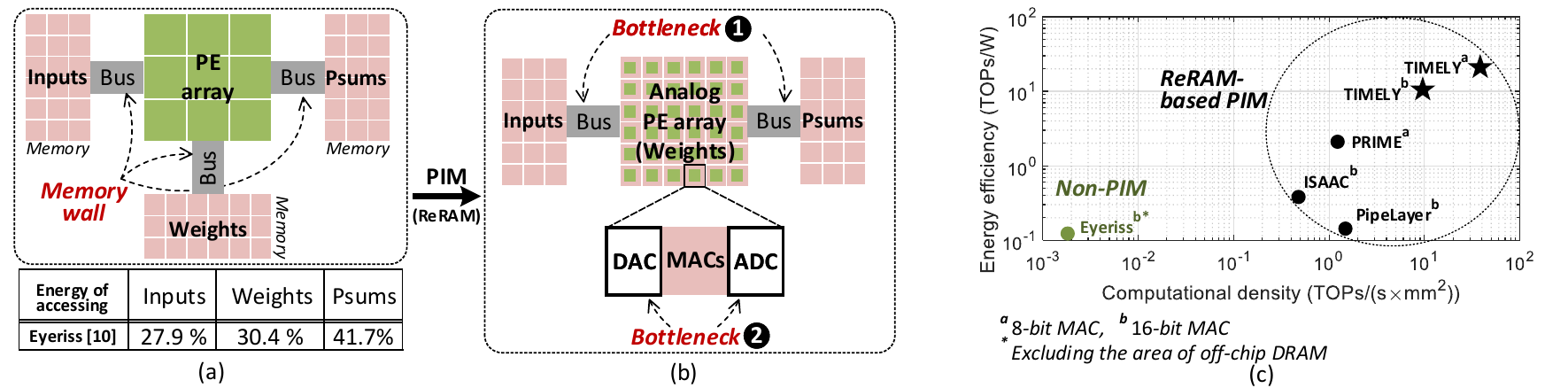}
	\vspace{-10pt}
	\caption{An illustration of (a) the ``memory wall'' in \textcolor[rgb]{0,0,0}{CNN/}DNN accelerators due to data movements of inputs, weights, and Psums, and an example of their energy breakdown~\cite{Eyeriss}, (b) the energy efficiency bottlenecks of PIM accelerators: (1) input and Psum movements (i.e. Bottleneck {{\color{black}{\ding{182}}}}) and (2) the DAC/ADC \textcolor[rgb]{0,0,0}{interfacing} (i.e. Bottleneck {{\color{black}{\ding{183}}}}), and (c) bench-marking the energy efficiency and computational density of the proposed \systemname over state-of-the-art \textcolor[rgb]{0,0,0}{CNN/}DNN accelerators, including a non-PIM accelerator (Eyeriss~\cite{Eyeriss}) and R$^2$PIM accelerators (PRIME~\cite{PRIME}, ISAAC~\cite{ISAAC}, and PipeLayer~\cite{PipeLayer}).}
	\label{bottlenecks_state_of_the_arts}
	\vspace{-10pt}
\end{figure*}

To close the gap between the constrained resources of IoT devices and the growing complexity of \textcolor[rgb]{0,0,0}{CNNs/}DNNs, many energy-efficient accelerators have been proposed \cite{Ziyun-ISSC2019,Ziyun-JSSC2019,Eyeriss,DianNao,7780065,Xu_2020,hotchips_arm,nvdla}.
As the energy cost of \textcolor[rgb]{0,0,0}{CNN/}DNN accelerators is dominated by memory accesses of inputs, weights and partial sums (Psums) (see Fig.~\ref{bottlenecks_state_of_the_arts} (a)) (e.g., up to 95\% in DianNao~\cite{DianNao}), processing-in-memory (PIM) accelerators have emerged as a promising solution
\textcolor[rgb]{0,0,0}{in which the computation is moved into the memory arrays and weight movements are eliminated}
(see Fig.~\ref{bottlenecks_state_of_the_arts} (b)). Among PIM accelerators on various memory technologies ~\cite{MRAM,Drisa,Scope,ISAAC,PRIME,PipeLayer,8465832,Tsinghua_Nature2020}, resistive-random-access-memory-(ReRAM)-based-PIM (R$^2$PIM) accelerators have gained extensive research interest due to \textcolor[rgb]{0,0,0}{ReRAM's} high density (e.g. 25$\times$--50$\times$ higher over SRAM~\cite{ReRAM_density_ShimengYu,6193402}).
However, the energy efficiency of R$^2$PIM accelerators \textcolor[rgb]{0,0,0}{(such as PRIME~\cite{PRIME}, ISAAC~\cite{ISAAC}, and PipeLayer~\cite{PipeLayer})} is still limited due to two bottlenecks (see Fig.~\ref{bottlenecks_state_of_the_arts} (b)): \textit{(1) although the weights are kept stationary in memory, the energy cost of data movements due to inputs and Psums is still large  (as high as 83\% in PRIME~\cite{PRIME}); (2) the energy of the interfacing circuits  (such as \textcolor[rgb]{0,0,0}{analog-to-digital converters} (ADCs)/\textcolor[rgb]{0,0,0}{digital-to-analog converters} (DACs)) is another limiting factor (as high as 61\% in ISAAC~\cite{ISAAC}).} 

To address the aforementioned energy bottlenecks, we analyze and identify opportunities for greatly enhancing the energy efficiency of R$^2$PIM accelerators (see Section~\ref{3Opportunity}), and develop three novel techniques that strive to push data movements and interfaces in PIM accelerators towards local and in \textcolor[rgb]{0,0,0}{time domain} (see Section~\ref{3innovation}). While these three techniques are in general effective for enhancing the energy efficiency of PIM accelerators, we evaluate them in a R$^2$PIM accelerator, and
demonstrate an improvement of energy efficiency by one order of magnitude \textcolor[rgb]{0,0,0}{over state-of-the-art R$^2$PIM accelerators}. The contribution of this paper is as follows:

\begin{itemize}
\item We propose three new ideas for aggressively improving energy efficiency of R$^2$PIM accelerators: (1) adopting analog local buffers (ALBs) within memory crossbars for enhancing (analog) data locality, (2) time-domain interfaces (TDIs) to reduce energy cost of single digital-to-analog (D/A) (and analog-to-digital (A/D)) conversion, and (3) a new mapping method called only-once input read (O$^2$IR) to further save the number of input/Psum accesses and D/A conversions.

\item We develop an innovative R$^2$PIM architecture (see Section IV), {\textbf{\systemname} (\textbf{T}ime-domain,  \textbf{I}n-\textbf{M}emory \textbf{E}xecution, \textbf{L}ocalit\textbf{Y})}, that integrates the three aforementioned ideas to (1) \ul{maximize (analog) data locality} via ALBs and O$^2$IR and (2) \ul{minimize the D/A (and A/D) interfaces' energy cost} by making use of the more energy-efficient TDIs\textcolor[rgb]{0,0,0}{, the ALBs}  and the O$^2$IR method. \systemname outperforms the most competitive R$^2$PIM accelerators in both energy efficiency (over PRIME) and computational density (over PipeLayer) (see Fig.~\ref{bottlenecks_state_of_the_arts} (c)). 

\item We perform a thorough evaluation of \systemname against \ul{4 state-of-the-art R$^2$PIM accelerators} on \ul{$>$10 CNN and DNN models} under various chip configurations, and show that \systemname~achieves up to 18.2$\times$ improvement (over \sloppy ISAAC) in energy efficiency, 31.2$\times$ improvement (over PRIME) in computational density, and 736.6$\times$ in throughput (over PRIME), demonstrating a promising architecture 
\textcolor[rgb]{0,0,0}{for accelerating \textcolor[rgb]{0,0,0}{CNNs and} DNNs}. Furthermore, we perform ablation studies to evaluate the effectiveness of each \systemname's feature (i.e., ALB, TDI, and O$^2$IR) in reducing energy and area costs, and demonstrate that TIMELY's innovative ideas can be generalized to other R$^2$PIM accelerators. 

\end{itemize}
\vspace{-5pt}
\section{Background}
This section provides the background of R$^2$PIM  CNN/DNN accelerators. First, we introduce CNNs  and the input reuse opportunities in CNNs' convolutional (CONV) operations in Section~\ref{CNN_InputReuse}, and ReRAM basics in Section~\ref{sec_reram_basics}. Second, we compare \textcolor[rgb]{0,0,0}{digital-to-time converter} (DTC)/\textcolor[rgb]{0,0,0}{time-to-digital converter} (TDC) and DAC/ADC, which are two types of digital-to-analog (D/A) and analog-to-digital (A/D) conversion, in terms of energy costs and accuracy in Section~\ref{DTC_vs_DAC}.

\vspace{-5pt}
\subsection{CNN and Input Reuse}
\label{CNN_InputReuse}

CNNs are composed of multiple CONV layers. \textcolor[rgb]{0,0,0}{Given the CNN parameters in Table~\ref{paras_TIMELY},} the computation in a CONV layer can be described as:
\begin{small}
\begin{equation}
\label{CNN}
\begin{split}
&O[v][u][x][y]=\sum_{k=0}^{C-1}{\sum_{i=0}^{G-1}{\sum_{j=0}^{Z-1}{I[v][k][Sx+i][Sy+j]\times W[u][k][i][j]}}}\\
&+B[u],~~~~~~~~~~~~~~0\le v < M, 0\le u < D, 0\le x < F, 0\le y < E\\
\end{split}
\end{equation}
\end{small}
where $O$, $I$, $W$, and $B$ denote matrices of the output feature maps, input feature maps, filters, and biases, respectively.
\textcolor[rgb]{0,0,0}{Fully-connected (FC) layers are typically behind CONV layers. Different from CONV layers, the filters of FC layers are of the same size as the input feature maps~\cite{Eyeriss}.} 
Equation (\ref{CNN}) can describe FC layers with additional constraints, i.e., $Z=H$, \textcolor[rgb]{0,0,0}{$G=W$}, $S=1$, and $E=F=1$. 

\begin{figure}[!b]
	\centering
	\vspace{-15pt}
	\includegraphics[width=90mm]{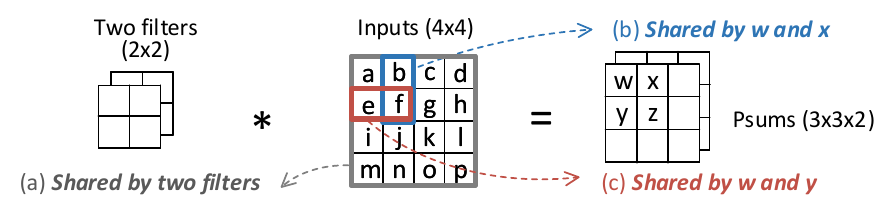}
	\vspace{-25pt}
	\caption{{Illustrating the three types of input reuses.}}
	\label{3D_reuse_opp}
 	\vspace{-10pt}
\end{figure}

Three types of input reuses exist in \textcolor[rgb]{0,0,0}{CNN} CONV operations yielding 3-D Psums. Consider the example in Fig.~\ref{3D_reuse_opp} where $C$ and $M$ are set to 1 for simplicity because input reuses are independent on them. First (see Fig.~\ref{3D_reuse_opp} (a)), one input feature map is shared by multiple (e.g. two in Fig.~\ref{3D_reuse_opp}) output channels' filters. Second (see Fig.~\ref{3D_reuse_opp} (b)), as filters slide horizontally, input pixels are reused to generate outputs in the same row \textcolor[rgb]{0,0,0}{-- e.g.} $b$ and $f$ are used twice to generate $w$ and $x$, respectively. Third (see Fig.~\ref{3D_reuse_opp} (c)), as filters slide vertically, input pixels are reused to generate outputs in the same column \textcolor[rgb]{0,0,0}{-- e.g.} $e$ and $f$ are used twice to generate $w$ and $y$, respectively. Given a layer with $D$ output channels, a filter size of $Z\times G$, and a stride of $S$, each input pixel is reused $DZG/S^2$ times~\cite{2018arXiv180604321Y}. For example, $f$ is reused 8 times in Fig.~\ref{3D_reuse_opp} where $D$=2, $Z$=$G$=2, and $S$=1. \textcolor[rgb]{0,0,0}{Note that weights are private in DNN CONV operations, which is the main difference between CNNs and DNNs~\cite{Cambricon-X}}.

\vspace{-5pt}
\subsection{ReRAM Basics}
\label{sec_reram_basics}
\begin{figure}[!t]
	\centering
	\vspace{-5pt}
	\includegraphics[width=90mm]{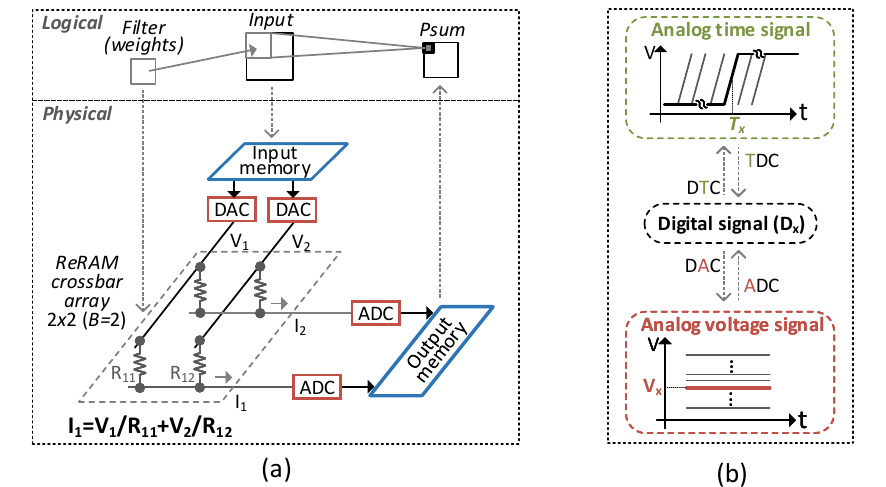}
	\vspace{-20pt}
	\caption{(a) ReRAM operation basics and (b) two types of interfacing circuits.}
\vspace{-15pt}
	\label{ReRAMbasics}
\end{figure}

ReRAM is a type of nonvolatile memory \textcolor[rgb]{0,0,0}{storing} data through resistance modulation~\cite{6193402,ReRAMtype1,ReRAMtype2,TSMC_ASSCC2017,Intel_ISSCC2019}. \textcolor[rgb]{0,0,0}{An ReRAM} cell with a metal-insulator-metal (MIM) structure consists of top/bottom electrodes and a metal-oxide layer~\cite{6193402}. Analog multiplication can be performed in ReRAM cells \textcolor[rgb]{0,0,0}{(see Fig.~\ref{ReRAMbasics} (a))}, with the biased voltages \textcolor[rgb]{0,0,0}{serving} as inputs, ReRAM cells' conductance as weights, and resulting currents as outputs. Addition operations are realized through current summing among ReRAM cells of the same columns \cite{hu2016dot, yao2017face} \textcolor[rgb]{0,0,0}{-- e.g.} $I_1=V_1/R_{11}+V_2/R_{12}$ in Fig.~\ref{ReRAMbasics} (a). At the circuit level, digital inputs are read from an input memory, converted to analog voltages by DACs, and then applied on ReRAM cells. The resulting analog Psums are converted to digital values by ADCs, and then stored back into an output memory.

\vspace{-20pt}
\begin{table}[b!]
\vspace{-10pt}
\def\arraystretch{1}
\centering
\caption{A summary of parameters used in \systemname}
\vspace{-5pt}
\scriptsize
\begin{tabular}{|c|c|}
\hline
{\textbf{CNN Params}} & {\textbf{Description}}  \\
\hline
\hline
{$M $} & {batch size of 3-D feature maps  }\\
    \hline
{$C / D$} & {input / output channel  }\\
    \hline
{$H / W $} & {input feature map height / width  } \\
    \hline
{$Z  / \textcolor[rgb]{0,0,0}{G} $} & {filter height / \textcolor[rgb]{0,0,0}{width}} \\
    \hline
{$ S $} & {stride } \\
    \hline
{$E  / F $} & {output feature map height / width  }\\
    \hline
    \hline
{\textbf{Arch. Params}} & {\textbf{Description}}  \\
    \hline
        \hline
$B$ &\# of ReRAM bit cells in one crossbar array is $B^2$\\
    \hline
$N_{CB}$ & {\# of ReRAM crossbar arrays in one sub-Chip is $N_{CB}^2$}\\
    \hline
\multirow{2}{*}{$R_{ij}$}  & the resistance of the ReRAM bit cell at the $i^{th}$ row \\
~  & and $j^{th}$ column of a crossbar array \\
    \hline
$T_{i}$  & the time input for the $i^{th}$ row of an ReRAM crossbar array\\
    \hline
$T_{o,8b/4b}$  &     the time Psum for 8-bit inputs and 4-bit weights\\
    \hline
$VDD$  & the logic high voltage of the time-domain signals\\
    \hline
$V_{th}$  &  the threshold voltage of a comparator \\
    \hline
$C_{c}$  &  the charging capacitance\\
    \hline
$T_{del}$ & the unit delay of a DTC/TDC\\
    \hline
\multirow{2}{*}{$\gamma$} &   {one} DTC/TDC is shared by $\gamma$ rows/columns\\  
~ &   in one ReRAM crossbar array\\
    \hline 
$\phi$ & the reset phase of a sub-Chip (reset: $\phi$=1)\\
\hline 
\textcolor[rgb]{0,0,0}{$\chi$} & \textcolor[rgb]{0,0,0}{the number of sub-Chips in one TIMELY chip}\\
\hline 
\textcolor[rgb]{0,0,0}{$\varepsilon$} &  \textcolor[rgb]{0,0,0}{the potential error of one X-subBuf}\\
\hline
{\textbf{Energy Params}} & {\textbf{Description}}  \\
\hline
\hline
$e_{DTC}$  &  the energy of one conversion in DTC\\
    \hline
$e_{TDC}$  &  the energy of one conversion in TDC\\
    \hline
    $e_{DAC}$   &  the energy  of one conversion in DAC\\
    \hline
$e_{ADC}$ &  the energy of one conversion in ADC\\
    \hline
$e_{P}$  &  the unit energy of accessing P-subBuf\\
    \hline
$e_{X}$  &  the unit energy of accessing X-subBuf\\
    \hline
$e_{R^2}$  &  the unit energy of accessing ReRAM {input/output buffers}\\
    \hline
\end{tabular}
\vspace{-5pt}
\label{paras_TIMELY}
\end{table}

\vspace{12pt}
\subsection{DTCs/TDCs vs. DACs/ADCs
\vspace{-2pt}
\label{DTC_vs_DAC}}
As shown in Fig.~\ref{ReRAMbasics} (b), DTCs/TDCs can perform the con-

\noindent version between an analog time signal and the corresponding digital signal; DACs/ADCs  can do so between an analog voltage signal and the digital signal.  
One digital signal (e.g. $D_x$ in Fig.~\ref{ReRAMbasics} (b)) can be represented as a time delay with a fixed high/low voltage (corresponding to 1/0) in the time domain (e.g. $T_x$ in Fig.~\ref{ReRAMbasics} (b))~\cite{timeinput1,timeinput2,timeinput3,timeinput4}, or as a voltage  {in} the voltage domain (e.g. $V_x$ in Fig.~\ref{ReRAMbasics} (b)).
Compared with a DTC/TDC which can be implemented using digital circuits~\cite{miyashita201310,amravati201855nm,TDC,zhang2017time,5308604, 6800123}, a DAC/ADC  {typically} relies on analog circuits that (1) are more power consuming and (2) vulnerable to noises and process, voltage and temperature (PVT) variations, and (3) benefit much less from  {process} scaling in energy efficiency~\cite{7993627}.

\section{Opportunities and Innovations}
This section aims to answer the question of \textbf{``\textcolor[rgb]{0,0,0}{how can} \systemname~outperform state-of-the-art R$^2$PIM accelerators?''} 
Note that all parameters used in this section are summarized in Table~\ref{paras_TIMELY}.
\subsection{Opportunities}
\label{3Opportunity}
We first identify three opportunities for greatly reducing energy costs of R$^2$PIM accelerators by analyzing performance limitations in state-of-the-art designs. Specifically, Opportunity \#1 is motivated by the energy bottleneck of (1) input and Psum movements (i.e., Bottleneck {{\color{black}{\ding{182}}} in  Fig.~\ref{bottlenecks_state_of_the_arts} (b)) and (2) interfacing circuits (i.e., Bottleneck {{\color{black}{\ding{183}}}} in Fig.~\ref{bottlenecks_state_of_the_arts} (b)); Opportunity \#2 is inspired by the bottleneck of interfacing circuits; and Opportunity \#3 is motivated by both types of bottlenecks.
\begin{figure}[!b]
	\centering
   \vspace{-10pt}
	\includegraphics[width=90mm]{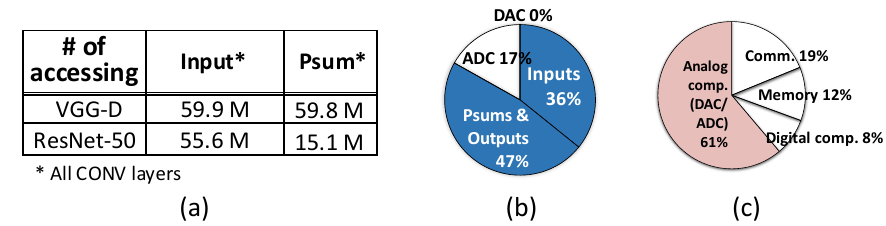}
	\vspace{-20pt}
	\caption{(a) The number of input/Psum accesses, (b) energy breakdown of PRIME~\cite{PRIME}, and (c) energy breakdown of ISAAC~\cite{ISAAC}.}
	\label{Energy_bottleneck_prior_work}
	\vspace{-10pt}
\end{figure}

\textbf{Opportunity \#1.} \ul{Enhancing (analog) data locality to greatly reduce the energy/time costs of both data movements and D/A and A/D interfaces.} We identify this opportunity based on the following considerations. Since in-ReRAM processing computes in the analog domain, the operands, including inputs, weights, and Psums, are all analog. If we can mostly access analog operands locally, we can expect large energy savings associated with input and Psum movements and largely  remove  the need to activate D/A and A/D interfaces. In the prior works, the input/Psum movements and interfaces dominate the energy cost of R$^2$PIM accelerators.} First, input and Psum accesses involve energy-hungry data movements.
While weights stay stationary in R$^2$PIM accelerators, input and Psum accesses are still needed. Although one input{/}Psum access can be shared by $B$ ReRAM cells in the same row{/}column, for dot-product operations in a $B\times B$ ReRAM crossbar array, a large number of input and Psum accesses are still required. For example, more than 55 million inputs and 15 million Psums need to be accessed during the VGG-D ~\cite{VGG} and ResNet-50~\cite{7780459} inferences, respectively (see Fig.~\ref{Energy_bottleneck_prior_work} (a)).
While inputs require only memory read, Psums involve both memory write and read, resulting in a large energy cost. As an example, 36\% and 47\% of the total energy in PRIME~\cite{PRIME} are spent on input and Psum accesses, respectively  {(see Fig.~\ref{Energy_bottleneck_prior_work} (b))}. Second, voltage-domain D/A and A/D conversions involve a large energy cost. For example, in PRIME, except the data movement energy, most of the remaining energy cost is consumed by D/A and A/D conversions (see Fig.~\ref{Energy_bottleneck_prior_work} (b)).

{\textbf{Opportunity \#2.} \ul{Time-domain interfacing can reduce the energy cost of a single D/A (and A/D) conversion.}} Since time-domain D/A and A/D conversion is more energy efficient than voltage-domain conversion (see Section~\ref{DTC_vs_DAC}), we have an opportunity to use DTCs and TDCs for interfacing between the digital signals stored in memory and analog signals computated in ReRAM crossbar arrays. In prior works, DACs and ADCs limit the energy efficiency of R$^2$PIM accelerators. Although ISAAC optimizes the energy cost of its DAC/ADC interface, the interface energy is still as large as 61\% in ISAAC (see Fig.~\ref{Energy_bottleneck_prior_work} (c)).
Specifically, ISAAC~\cite{ISAAC} decreases the number of ADCs by sharing one ADC among 128 ReRAM bitlines, and thus the ADC sampling rate increases by 128$\times$, increasing the energy cost of each A/D conversion.

{\textbf{Opportunity \#3.} \ul{Reducing the number of input accesses can save the energy cost of both input accesses and D/A conversions.} We find that the input reuse of CNNs can still be improved over the prior works for reducing the energy overhead of input accesses and corresponding interfaces.
Though each input connected to one row of an ReRAM array is naturally shared by $B$ ReRAM cells along the row, each input on average has to be accessed $DZG/S^2/B$ times. Taking ISAAC~\cite{ISAAC} as an example, one 16-bit input involves $DZG/S^2/B$ times unit eDRAM read energy (i.e. 4416$\times$  {the energy} of a 16-bit ReRAM MAC), input register file read energy  (i.e. 264.5$\times$  {the energy} of a 16-bit ReRAM MAC) and D/A conversion energy  (i.e. 109.7$\times$  {the energy} of 16-bit ReRAM MAC). For MSRA-3~\cite{msra} adopted by ISAAC, each input of CONV layers is read and activated the interfaces 47 times on average.
}

\vspace{-5pt}
\subsection{TIMELY Innovations}
\label{3innovation}
\begin{figure}[!t]
	\centering
	\vspace{-15pt}
	\includegraphics[width=86mm]{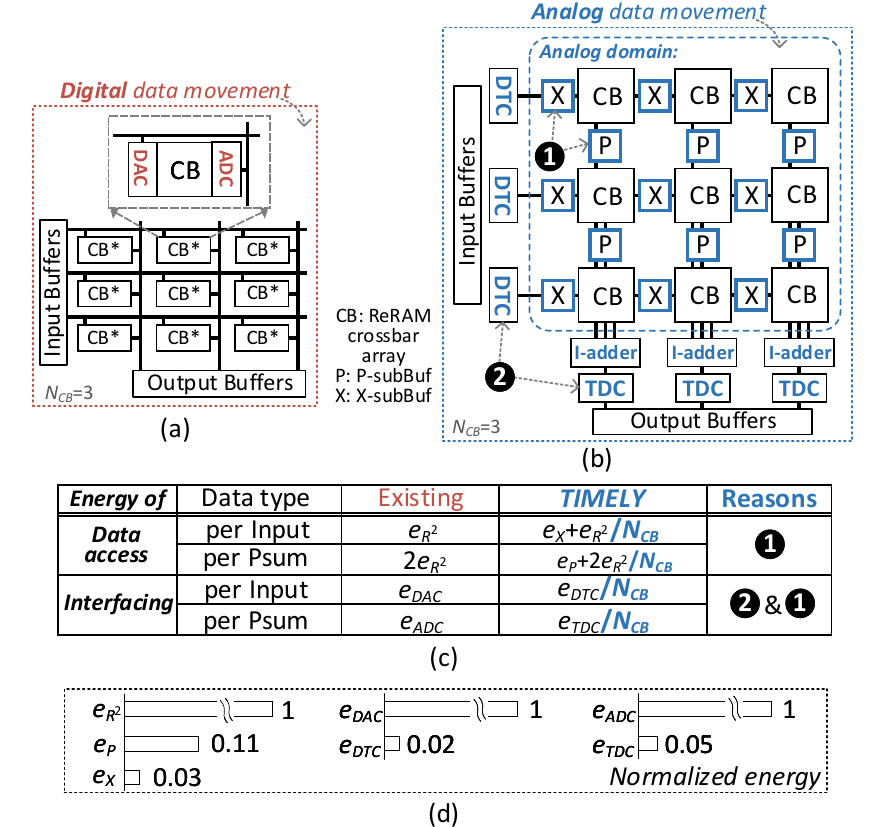}
	\vspace{-10pt}
	\caption{ A high-level view of (a) a sub-Chip within a chip of state-of-the-art R$^2$PIMs and (b) \systemname's sub-Chip, (c) the energy cost per input and per Psum   {in} state-of-the-art R$^2$PIMs and \systemname, and (d) the normalized energy of different data accesses and interfaces, where $e_{R^2}$, $e_X$, and $e_P$ are the unit energy of accessing ReRAM   {input/output buffers}, X-subBuf, and P-subBuf, respectively, while $e_{DAC}$, $e_{ADC}$, $e_{DTC}$, and $e_{TDC}$ denote the energy of one DAC, ADC, DTC, and TDC \cite{PRIME,TDC,7765065,8bDAC,SARadc,ISAAC}, respectively.
	}
	\label{TIME-innovation}
	\vspace{-15pt}
\end{figure}

The three aforementioned opportunities inspire us to develop the three innovations in \systemname for greatly improving the acceleration energy efficiency.
Fig.~\ref{TIME-innovation} (a) and (b) show a conceptual view of the difference between existing R$^2$PIM accelerators and \systemname. Specifically, \systemname mostly moves data in the analog domain as compared to the fully digital data movements in the existing designs and adopts DTCs and TDCs instead of DACs and ADCs for interfacing. 

\textbf{Innovation \#1.} \ul{TIMELY adopts ALBs to aggressively enhance (analog) data locality,} leading to about $N_{CB}\times$ reduction in data movement energy costs per input and per Psum compared with existing designs, assuming a total of $N_{CB} \times N_{CB}$ crossbars in each sub-Chip. Multiple sub-Chips compose one chip.
One key difference between \systemname~and existing R$^2$PIM resides in their sub-Chip design (see Fig.~\ref{TIME-innovation} (a) vs. (b)). Specifically, each crossbar (i.e. CB in Fig.~\ref{TIME-innovation}) in \ul{existing designs} fetches inputs from a high-cost memory (e.g. input buffers in Fig.~\ref{TIME-innovation} (a)). Therefore, for each sub-Chip, there is an energy cost of $BN_{CB}^2e_{R^2}$ for accessing $BN_{CB}^2$ inputs. In \ul{\systemname} (see Fig.~\ref{TIME-innovation} (b)), an input fetched from the high-cost memory is shared by one row of  the sub-Chip thanks to the adopted local ALB buffers  {(e.g. X-subBufs in Fig.~\ref{TIME-innovation} (b))} that {are} sandwiched  {between} the crossbar arrays, {resulting in an energy cost of $BN_{CB}e_{R^2}+BN_{CB}^2e_X$ for handling the same number of inputs, leading to an energy reduction of  {$N_{CB}\times$} per input (see Fig.~\ref{TIME-innovation} (c).} Similarly, each crossbar in existing R$^2$PIM accelerators directly writes and  {reads} Psums to and from the high-cost  {output buffers}, whereas in {\systemname} the Psums in each column of the  {sub-Chip} are accumulated before being written back to the {output buffers, leading to an energy cost reduction of  {$N_{CB}\times$} per Psum (see Fig.~\ref{TIME-innovation} (c).}
 {Furthermore,} accessing the high-cost memory requires about one order of \textcolor[rgb]{0,0,0}{magnitude} higher energy cost than that of a local buffer. Specifically,  {the average energy of} one high-cost memory access in PRIME is about  {$9\times$} and  {$33\times$} higher than that of P-subBufs and X-subBufs ~\cite{ReRAM_density_ShimengYu} in \systemname, respectively. $N_{CB}$ is typically $>$10 (e.g. $N_{CB} = 12$ in PRIME).  {Therefore, about} $N_{CB}\times$ energy reduction for handling {input/Psum accesses can be achieved in} \systemname.   {Additionally, the much reduced requirements of input/output buffer size in \systemname make it possible to eliminate inter sub-Chip memory (see Fig.~\ref{time_comp_storage} (a) and Fig.~\ref{TIME_PRIME_interface} (c), leading to additional energy savings.}

\textbf{Innovation \#2.} \ul{TIMELY adopts TDIs and ALBs to minimize the energy cost of a single conversion and the total number of conversions, respectively.} As a result, \systemname reduces the  {interfacing} energy cost per input and per Psum by $q_1N_{CB}$ and $q_2 N_{CB}$, respectively, compared with current practices, where \sloppy $q_1=e_{DAC}/e_{DTC}$ and $q_2=e_{ADC}/e_{TDC}$.
It is well recognized that the energy cost of ADC/DAC interfaces is  {another bottleneck} in existing  {R$^2$PIM} accelerators {, in addition to} that of data movements. For example, the energy cost of ADCs  {and} DACs in ISAAC accounts for $>$61\% of its total energy cost. In contrast, \systemname adopts (1) TDCs/DTCs instead of ADCs/DACs to implement the interfacing circuits of crossbars and (2) only one TDC/DTC conversion for each row/column of \textbf{one sub-Chip}, whereas each  {row/column of} \textbf{crossbar} {needs} one ADC/DAC conversion in existing designs, leading to a total of $q_1N_{CB}\times$ and $q_2 N_{CB}\times$ reduction per input and Psum, respectively, as compared to existing designs. Specifically,  {$q_1$ and $q_2$ are about 50 and 20}~\cite{TDC,7765065,8bDAC,SARadc,ISAAC}, respectively.

\textbf{Innovation \#3.} \ul{TIMELY employs O$^2$IR to further reduce the number and thus energy cost of input accesses and D/A conversions.}  As accessing the input and output buffers in sub-Chips costs about one order of magnitude higher energy than that of accessing local buffers    {between} the crossbar arrays    {(see the left part of Fig.~\ref{TIME-innovation} (d))}, we propose an O$^2$IR strategy to increase the input reuse opportunities for minimizing the cost of input accesses and associated D/A conversion.

\section{TIMELY Architecture}
\label{sec:proposed}

In this section, \textcolor[rgb]{0,0,0} {we first show an architecture overview (see Section IV-A), and then describe how the \systemname~architecture integrates the three innovations for aggressively improving the acceleration energy efficiency in Sections~\ref{TIMELYanalogmove},~\ref{dot_product_time}, and~\ref{OOIR}, respectively. In addition, we introduce our pipeline design for enhancing throughput in Section~\ref{pipeline_arch} and the software-hardware interface design for offering programmability in Section~\ref{interface_arch}. 
Parameters are summarized in Table~\ref{paras_TIMELY}.}

\begin{figure}[!t]
	\centering
	\vspace{-10pt}
	\includegraphics[width=86mm]{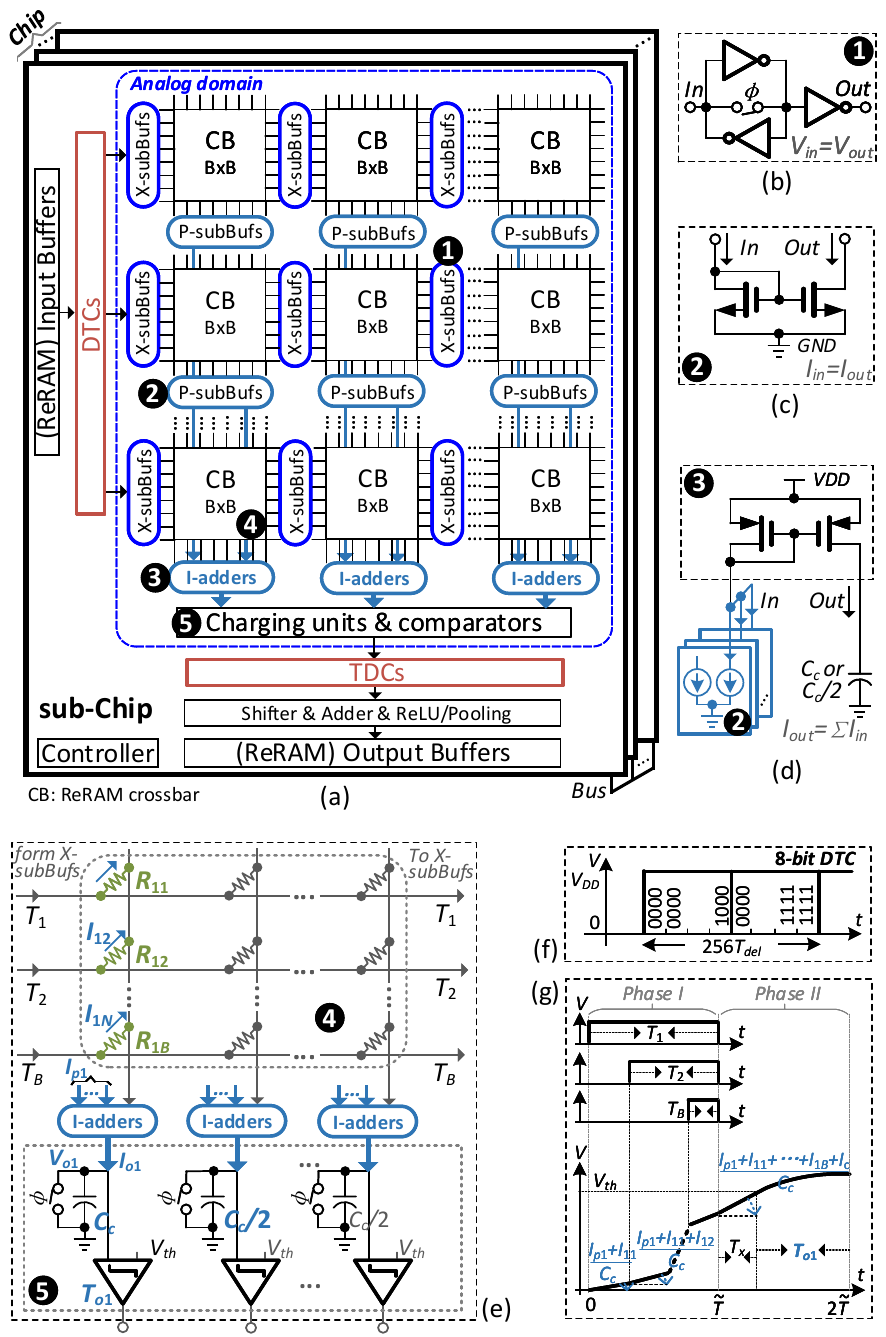}
	\vspace{-5pt}
	\caption{An illustration  of the (a) TIME{LY} architecture: (b) {\color{black}{\ding{182}}} X-subBuf, (c) {\color{black}{\ding{183}}} P-subBuf, (d) {{{\color{black}{\ding{184}}} I-adder}, (e) {\color{black}{\ding{185}}} ReRAM crossbar located in the first crossbar column and last row of a sub-Chip, {\color{black}{\ding{186}}} charging units, and comparators, (f) the input/output characteristics of an 8-bit DTC, and (g) {the input/output characteristic of dot-product operations} in the leftmost {ReRAM} column of a sub-Chip}.}
	\vspace{-15pt}
	\label{time_comp_storage}
\end{figure}

{\subsection{Overview}}
Fig. \ref{time_comp_storage} (a) shows the \systemname~architecture, which consists of a number of sub-Chips connected via bus \cite{PRIME,ISAAC}. Specifically, each sub-Chip includes DTCs/TDCs (on the left/at the bottom), ReRAM input/output buffers (on the left/at the bottom), ReRAM crossbars \textcolor[rgb]{0,0,0}{(see {{\color{black}{\ding{185}}}} in Fig.~\ref{time_comp_storage} (a))} with each having   {$B \times B$} bit cells, a mesh grid of local ALB buffers -- i.e., X-subBufs \textcolor[rgb]{0,0,0}{(see {{\color{black}{\ding{182}}}} in Fig.~\ref{time_comp_storage} (a))} and P-subBufs \textcolor[rgb]{0,0,0}{(see {{\color{black}{\ding{183}}}} in Fig.~\ref{time_comp_storage} (a))} -- {between} the ReRAM crossbar arrays,   {current adders (i.e. I-adders, \textcolor[rgb]{0,0,0}{{\color{black}{\ding{184}}} in Fig.~\ref{time_comp_storage} (a))}}, and a block of shift-and-add, \textcolor[rgb]{0,0,0}{ReLU}, max-pooling units. 

The \systemname architecture processes {CNNs/}DNNs' inference as follows. The pre-trained weights are pre-loaded into \systemname's ReRAM arrays. Inputs of the CNN/DNN layers are fetched into the input buffers of one sub-Chip or several sub-Chips that handle the corresponding layers, starting from the first CNN/DNN layer.
Within each sub-Chip, the inputs are applied to the DTCs for converting the digital inputs into analog time signals, which are then shared by ReRAM bit cells in the same row of all crossbar {arrays} along the horizontal direction to perform dot products with the corresponding resistive weights. \textcolor[rgb]{0,0,0}{The calculated Psums at the same column of all crossbars in the vertical direction} are aggregated in the I-adders, and converted into a voltage signal and then an analog time signal \textcolor[rgb]{0,0,0}{by a charging unit and comparator block (see {{\color{black}{\ding{186}}}} in Fig.~\ref{time_comp_storage} (a))} before being converted into a digital signal via a TDC. \textcolor[rgb]{0,0,0}{Note that the output of each P-subBuf is connected to the I-adder separately.} Finally, the resulting digital signals are applied to the block of shift-and-add, \textcolor[rgb]{0,0,0}{ReLU}, max-pooling units, and then written to the output buffers.
\vspace{-5pt}
\subsection{Enhancing (Analog) Data Locality}\label{TIMELYanalogmove}
\textcolor[rgb]{0,0,0}{Within each sub-chip of \systemname, the converted inputs and calculated Psums are moved in the analog domain with the aid of the adopted ALBs (see Fig.~\ref{time_comp_storage} (a)) after the digital inputs are converted into time signals by DTCs and before the Psums are converted into digital signals by TDCs. In this subsection, we first introduce the data movement mechanism and then present the operation of the local analog buffers.}  

\textbf{Data Movement Mechanism.} In the \systemname architecture, 
time inputs from the DTCs move horizontally across the ReRAM crossbar {arrays} in the same row via X-subBufs (see {{\color{black}{\ding{182}}}} in Fig.~\ref{time_comp_storage} (a))) for maximizing input reuses and minimizing  high-cost memory accesses. Meanwhile, the resulting current Psums move vertically via P-subBufs (see {{\color{black}{\ding{183}}}} in Fig.~\ref{time_comp_storage} (a)). 
Note that only the crossbars in the leftmost column fetch inputs from DTCs while those in all the remaining columns fetch inputs from their analog local time buffers (i.e., the X-subBufs to their left). Similarly, only the outputs of the I-adders are converted into the digital signals via TDCs before they are stored back into the output buffers, while the current outputs of the crossbars are passed into the I-adders via analog current buffers (i.e., the P-subBufs right below them). In this way, \systemname~processes most data movements in the analog domain within each sub-chip, greatly enhancing data locality for improving the energy efficiency and throughput. 

\textbf {Local Analog Buffers.} The local analog buffers make it possible to handle most (analog) data movements locally in \systemname. Specifically, X-subBuf buffers the time signals (i.e., outputs of the DTCs) by latching it, i.e., copying the input delay time to the latch outputs (see Fig.~\ref{time_comp_storage} (b)); while P-subBuf buffers the current signal outputted from the ReRAM crossbar array, i.e. copying the input current to their outputs (see Fig.~\ref{time_comp_storage} (c)). 
The key is that
X-subBuf and P-subBuf  are more energy and area efficient than input/output buffers ({see Fig.~\ref{TIME-innovation} (d)}). Specifically, an X-subBuf buffer consists of two cross-coupled inverters that form a positive feedback to speed up the response at its output and thus reduces the delay between its inputs and outputs~\cite{ADCbook}. Since cross-coupled inverters invert the input, a third inverter is used to invert the signal back. X-subBufs are reset in each  {pipeline-}cycle   {by setting $\phi$ to be high (see Fig.~\ref{time_comp_storage} (b)).} The P-subBuf buffer is implemented using an \textcolor[rgb]{0,0,0}{NMOS-pair} current mirror (see Fig.~\ref{time_comp_storage} (c))~\cite{Currentmirror1}. 

\subsection{Time-Domain Dot Products and DTC/TDC Interfacing}\label{dot_product_time}
\textcolor[rgb]{0,0,0}{\systemname performs dot products with time-domain inputs from the DTCs and converts time-domain dot product results into digital signals via TDCs. In this subsection, we first present dot product operations in TIMELY and then introduce their associated DTCs/TDCs.} 

\textcolor[rgb]{ 0,0,0}{\textbf{Dot Products.} 
\textcolor[rgb]{0,0,0}{First, let us consider Psums in one ReRAM crossbar array. 
Take the first column of the ReRAM crossbar array in Fig. \ref{time_comp_storage} (e) as an example.} A total of $B$ time-domain inputs $T_i~(i=1,2,...,B)$ are applied to their corresponding ReRAM bit cells with resistance values of $R_{1i}$ (i.e. corresponding to weights) to generate a Psum current \textcolor[rgb]{ 0,0,0}{(i.e. $T_i$-controlled current)} based on the Kirchoff's Law. 
Then, let us focus on Psums in one sub-Chip.} 
The Psum currents at the same column of all $N_{CB}$ crossbars \textcolor[rgb]{0,0,0}{in the vertical direction} are aggregated in the {I-adder~\cite{kaur2014current}} {(see {{\color{black}{\ding{184}}}} in Fig.~\ref{time_comp_storage} (a))}, 
and then are converted into a voltage $V_{o1}$ by charging a capacitor (e.g. $C_c$ \textcolor[rgb]{0,0,0}{in Fig.~\ref{time_comp_storage} (e)}). 
Fig.~\ref{time_comp_storage} ({g}) shows the input/output characteristic of the dot product. We adopt a 2-phase charging scheme~\cite{timeinput4}. In phase I, the charging time is the input $T_i$ and the charging current is ${VDD}/R_{1i}$, which corresponds to the weight. In phase II, the charging time is $T_x$ and the charging current is a constant $I_c$, which is equal to $C_cBN_{CB}V_{DD}/R_{min}$. The charging in phase II ensures the voltage on $C_c$ is larger than $V_{th}$, and the time output is defined by $\widetilde{T}-T_x$. $R_{min}$ is the minimum mapped resistance of one layer. 
$V_{th}$ is the threshold voltage of the comparator, which is equal to $BN_{CB}\widetilde{T}{V_{DD}}/{R_{min}}$, where $V_{DD}$ is the logic high voltage of the time signal, and $\widetilde{T}$ is the time period of one phase.
Based on Charge Conservation, we can derive the output $T_{o,8b/4b}$   {(see $T_{o1}$ in Fig.~\ref{time_comp_storage} (e))}, where 8b/4b represents 8-bit inputs and 4-bit weights, to be:

\vspace{-4pt}
\begin{small}
\begin{equation}
\label{}
T_{o,8b/4b}=\frac{R_{min}}{C_cBN_{CB}} \sum_{i=1}^{BN_{CB}}T_i/R_{1i}
\end{equation}
\end{small}
\vspace{-3pt}

%

To realize dot products with 8-bit weights and inputs, we employ a sub-ranging design~\cite{sar1,sar2,sar3} in which 8-bit weights are mapped into two adjacent bit-cell columns with the top-4 most significant bit (MSB) weights and the remaining 4 least significant bit (LSB) weights, respectively. The charging capacitors associated with MSB-weight column and LSB-weight column are $C_c$ and $C_c/2$, respectively. $T_{o,8b/4b}$ of the MSB-weight column and the LSB-weight column are added to get the dot-product result for 8-bit weights.

\textbf{DTCs/TDCs.}
We adopt 8-bit DTCs/TDCs for \systemname based on the measurement-validated designs in~\cite{TDC,7765065}. The input/output characteristics of a 8-bit DTC is shown in Fig. \ref{time_comp_storage} ({f}), where digital signals of ``1111111'' and ``00000000'' correspond to the time-domain analog signals with the maximum and minimum delays, respectively, and the dynamic range of the time-domain analog signals are $256 \times T_{del}$ with $T_{del}$ being the unit delay. Meanwhile, a TDC's input/output characteristics can also be viewed in Fig. \ref{time_comp_storage} ({f}) by switching the $V$ and $t$ axes. In \systemname, $T_{del}$ is designed to be 50 ps, leading to a conversion time of 25 ns \textcolor[rgb]{0,0,0}{(including a design margin)} for the 8-bit DTC/TDC.
In addition, to trade off energy efficiency and computational density,   {one} DTC/TDC is shared by   {$\gamma$ ($\gamma$ $\ge$1)} ReRAM crossbar rows/columns.
\subsection{\systemname's Only-Once Input Read Mapping Method}
\label{OOIR}
\begin{figure}[!t]
	\centering
    \vspace{-0pt}
	\includegraphics[width=86mm]{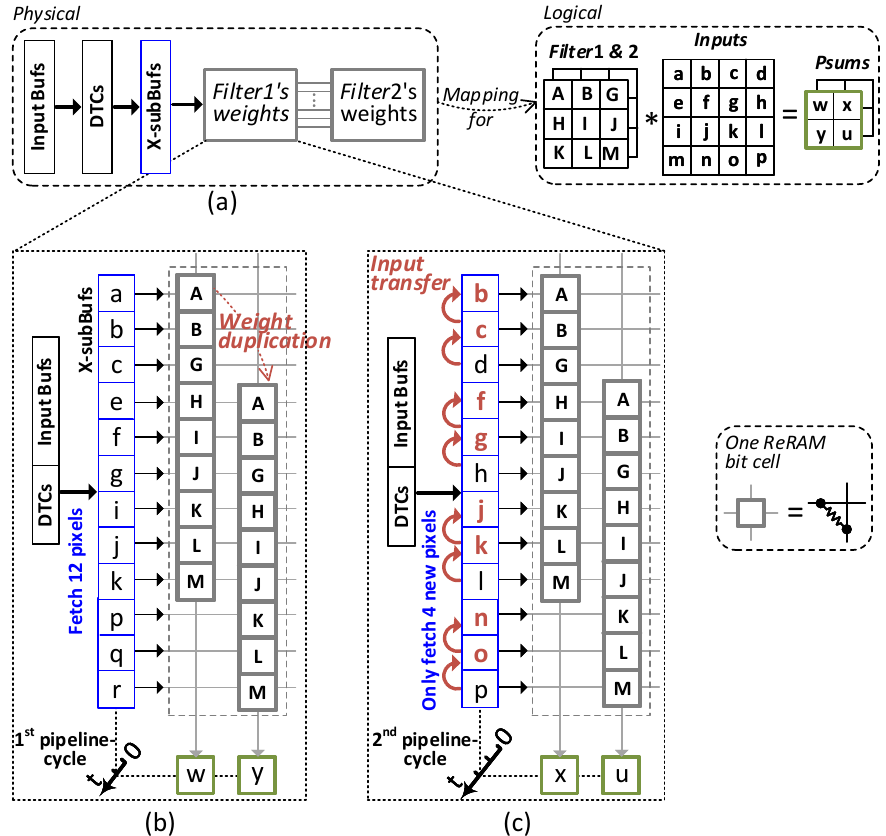}
	\vspace{-10pt}
	\caption{{The proposed O$^2$IR: (a) mapping filters using the same inputs into the same rows of crossbars; (b) duplicating filters with a vertical offset of $Z\times S$ between adjacent   {ReRAM} columns; and (c) temporally shifting inputs by an amount equal to $S$.}} 
	\vspace{-10pt}
	\label{3D_w_repli}
\end{figure}

O$^2$IR follows three principles: (1) for reusing the inputs by different filters, we map these filters   {in parallel within} the crossbar arrays (see Fig.~\ref{3D_w_repli} (a)); (2) for reusing the inputs when sliding the filter vertically within an input feature map, we duplicate the filters with a shifted offset equal to $Z\times S$ (see Fig.~\ref{3D_w_repli} (b)), where $Z$ and $S$ are the filter height and the stride, respectively; and (3) for reusing the inputs when sliding the filter horizontally within an input feature map, we transfer inputs to the adjacent X-subBufs with an step equal to $S$ (see Fig.~\ref{3D_w_repli} (c)). \textcolor[rgb]{ 0,0,0}{Single-direction input transfer between adjacent X-subBufs can be implemented by introducing only one switch and one control signal to one X-subBuf.}
\vspace{-0pt}
\subsection{Pipeline Design}
\label{pipeline_arch}
To enhance throughput, we adopt pipeline designs \textcolor[rgb]{ 0,0,0}{between and within sub-Chips, 
i.e., inter-sub-Chip and intra-sub-Chip pipeline. Different sub-Chips work in a pipeline way.} \textcolor[rgb]{ 0,0,0}{Note that a layer by layer weight mapping strategy is adopted in \systemname, where one CNN/DNN layer is mapped into one sub-Chip if the ReRAM crossbars' size is larger than the required size; otherwise, a layer is mapped into multiple sub-Chips.}
In one sub-Chip, {the following operations --} reading inputs from input buffers, DTCs, analog-domain computation (including dot-product, charging-and-comparison operations), TDCs, and writing back to output buffers   {-- are pipelined}.  
{The pipeline-cycle time is determined by the slowest stage.}
\textcolor[rgb]{0,0,0}{Let us take the operations within one sub-Chip as an example to illustrate the pipeline in \systemname. Assuming the first data is read from an input buffer at the first cycle, it spends three cycles to complete the digital-to-time conversion, analog-domain computation, and time-to-digital conversion, and is written back to an output buffer at the fifth cycle. Meanwhile, at the fifth cycle, the fifth, fourth, third, and second data is read, converted by a DTC, computed in the analog-domain, and converted by a TDC, respectively.  
}
\vspace{-5pt}
\subsection{Software-Hardware Interface} \label{interface_arch}
A software-hardware interface is adopted to allow developers to configure TIMELY for different {CNNs/DNNs}, enabling programmability. Similar to the interface in PRIME, three stages are involved from software programming to hardware execution. First, the {CNN/DNN} is loaded into an NN parser~\cite{zhang2018dnnbuilder} that automatically extracts model parameters. Second, with the extracted parameters, a compiler optimizes mapping strategies for increasing \textcolor[rgb]{0,0,0}{the utilization of ReRAM crossbar arrays} and then generates execution commands \textcolor[rgb]{0,0,0}{(including commands for weight mapping and input data path configuration)}.  Third, the controller (see Fig. \ref{time_comp_storage} (a)) loads the commands from the compiler to (1) write pre-trained weights to the mapped addresses, and (2) configure peripheral circuits for setting up \textcolor[rgb]{0,0,0}{input} paths of computation.

\vspace{-10pt}
\textcolor[rgb]{0,0,0}{{\section{Discussion}
\label{analog_discussion}}}

{Although local buffers have been adopted in {digital accelerators}{~\cite{Eyeriss,Cambricon-X}}, it is challenging when using local buffers in R$^2$PIMs because: (1) improper design can largely compromise R$^2$PIMs' high computational density and (2) more frequent large-overhead A/D and D/A conversions may be caused. 
To the best of our knowledge, TIMELY is the first to implement and maximize analog data locality via ALBs, which have at least one order of magnitude lower access energy cost compared to the two level memories in {PRIME~\cite{PRIME}/ISAAC~\cite{ISAAC}/Pipelayer~\cite{PipeLayer}}.  
Additionally, \systemname maximizes data locality without degrading R$^2$PIMs' computational density.}
\textcolor[rgb]{0,0,0}{Although a recent R$^2$PIM accelerator, CASCADE~\cite{CASCADE}, has adopted analog buffers, it only uses analog ReRAM buffer to reduce the number of A/D conversions, thereby minimizing computational energy. \systemname uses ALBs to minimize both the computational energy and data movement energy. Taking PRIME as an example, the computational energy only accounts for 17\% of the chip energy. In order to minimize the computational energy, \systemname not only reduces the number of A/D conversions by ALBs, but also decreases the energy of each A/D conversion by TDCs.}

\textcolor[rgb]{0,0,0}{Analog computations and local buffers are efficient, but they potentially introduce accuracy loss to \systemname. The accuracy loss is mainly attributed to the non-ideal characteristics of analog circuits. To address this challenge, \systemname not only leverages algorithm resilience of CNNs/DNNs to counter hardware vulnerability~\cite{Retraining_ReRAM_variation,Retraining_Reram_bit_failure,Retrain_JSSC17}, but also minimize potential errors introduced by hardware, thereby achieving the optimal trade-off between energy efficiency and accuracy.  \underline{First}, we choose time and current signals to minimize potential errors. Compared with analog voltage signals, analog current signals and digitally implemented time signals can tolerate larger errors caused by their loads, and analog time signal is less sensitive to noise and PVT variations~\cite{7993627}. \underline{Second}, the adopted ALBs help improve the accuracy of time inputs and Psums by increasing the driving ability of loads. However, the larger the number of ALBs, the smaller the number of ReRAM crossbar arrays in a sub-Chip, compromising the computational density. Based on system-level evaluations, we adopt one X-subBuf between each pair of neighboring ReRAM crossbar arrays and one P-subBuf between each ReRAM crossbar array and its I-adder in order to achieve a good trade-off between accuracy loss and computational density reduction. \underline{Third}, we limit the number of cascaded X-subBufs in the horizontal direction to reduce the accumulated errors (including noise) of time-domain inputs, which can be tolerated by the given design margin. We assign a design margin (i.e. more than 40 ps) for the unit delay (i.e. 50 ps) of the DTC conversion. We do not cascade P-subBufs to avoid introducing errors in Psum.}

\systemname~adopts pipeline designs to address the speed limit of time signal operations and thus improve throughput. Adjusting the number of ReRAM rows/columns shared by one DTC/TDC allows for the trade-off between the throughput and computational density of \systemname.
\textcolor[rgb]{0,0,0}{\systemname~compensates for the increased area due to the special shifted weight duplication of O$^2$IR (see Fig.~\ref{3D_w_repli} (b) and (c)) by saving peripheral circuits' area. Besides, 
\systemname~also replicates weights to improve computation parallelism and thus throughput, similar to prior designs~\cite{PRIME,ISAAC,PipeLayer}.}

\vspace{4pt}
\section{Evaluation}
\label{sec:experiment}

\textcolor[rgb]{0,0,0}{In this section}, we first introduce the experimental setup, and then compare \systemname~with state-of-the-art designs in terms of energy efficiency, computational density, and throughput. \textcolor[rgb]{0,0,0}{After that}, we demonstrate the effectiveness of \systemname's key features: \textcolor[rgb]{0,0,0}{ALB}, TDI, and O$^2$IR, and show that these features are generalizable. \textcolor[rgb]{0,0,0}{Finally, we discuss area scaling.}

\vspace{-5pt}
\subsection{Experiment Setup}

\textbf{\textcolor[rgb]{0,0,0}{\systemname} Configuration.}
{For a fair comparison with PRIME/ISAAC, we adopt PRIME/ISAAC's parameters, including ReRAM \textcolor[rgb]{0,0,0}{and ReLU} parameters from PRIME~\cite{PRIME}, and maxpool operations (scaled up to 65nm) and HyperTransport links from ISAAC~\cite{ISAAC} (see Table~\ref{TIMEhardware}). For \systemname's specific components, we use silicon-verified results~\cite{TDC,7765065} for DTCs and TDCs, and adopt Cadence-simulated results for X-subBuf, P-subBuf, I-adder, charging circuit, and comparator based on \cite{ADCbook, Currentmirror1, currentadder} {-- including their drives and loads during simulation.}} \textcolor[rgb]{0,0,0}{Supporting digital units (shifter and adder) consume negligibly small amounts of area and energy.} All the design parameters of the peripheral circuits are based on a commercial 65nm CMOS process. {The power supply is 1.2 V, and the clock rate is 40 MHz.} {The reset phase $\phi$ in Fig.~\ref{time_comp_storage} is 25 ns.}
{The pipeline-cycle time is determined by the latency of 8 (setting $\gamma$ to 8) DTCs/TDCs,} which have a larger latency than other pipelined operations. {The latency of reading corresponding inputs, analog-domain computations, and writing outputs back to output buffers are} 16 ns~\cite{ReRAMStanford}, 150 ns~\cite{ReRAMStanford}, and 160 ns~\cite{ReRAMStanford}{, respectively}.
{In addition, I-adders \textcolor[rgb]{0,0,0}{and its inputs} do not contribute to the total area because \textcolor[rgb]{0,0,0}{we insert} I-adders \textcolor[rgb]{0,0,0}{and the interconnection between each P-subBuf and I-adder} under the charging capacitors and ReRAM crossbars, leveraging different IC layers.} \textcolor[rgb]{0,0,0}{We adopt 106 sub-Chips in the experiments for a fair comparison with the baselines (e.g., TIMELY vs. ISAAC: 91mm$^2$ vs. 88 mm$^2$). }

\vspace{-0pt}
\begin{table}[t!]
\def\arraystretch{1}
\centering
\caption{\systemname~parameters.}
\vspace{-5pt}
\scriptsize
\begin{tabular}{|p{1.5cm}<{\centering}|p{1.3cm}<{\centering}|p{1.5cm}<{\centering}|p{1cm}<{\centering}|p{1cm}<{\centering}|}
\hline
\multirow{3}{*}{\textbf{Component}} & \multirow{3}{*}{\textbf{Params}} & \multirow{3}{*}{\textbf{Spec}} & \textbf{Energy ($fJ$)} & \textbf{Area ($\mu m^2$)} \\
~ & ~ & ~ & /compo. & /compo. \\
\hline
\hline
\multicolumn{5}{|c|}{\textbf{\systemname~sub-Chip}}\\
\hline
\multirow{2}{*}{DTC} & resolution & 8 bits & \multirow{2}{*}{37.5}  & \multirow{2}{*}{\textcolor[rgb]{0,0,0}{240}} \\
   & number & 16$\times$32  &  & \\ 
\hline
 ReRAM & size & 256$\times$256 & \multirow{3}{*}{1792} & \multirow{3}{*}{100}\\ 
 crossbar  & number &16$\times$12 &  &~\\
 ~     & bits/cell & 4 &  & \\ 
\hline
Charging+ & \multirow{2}{*}{number}&\multirow{2}{*}{\textcolor[rgb]{0,0,0}{12}$\times$256}&\multirow{2}{*}{41.7}&\multirow{2}{*}{\textcolor[rgb]{0,0,0}{40}}\\ 
comparator& & &   &~\\ 
\hline
\multirow{2}{*}{TDC} &  resolution   & 8 bits & \multirow{2}{*}{145} & \multirow{2}{*}{\textcolor[rgb]{0,0,0}{310}}\\
   & number     & 12$\times$32 &  &~\\ 
\hline
\multirow{1}{*}{X-subBuf} 
  & number & \textcolor[rgb]{0,0,0}{12}$\times$16$\times$256& 0.62 &\textcolor[rgb]{0,0,0}{5}\\ 
\hline
\multirow{1}{*}{P-subBuf} 
& number & 15$\times$12$\times$256& 2.3 &\textcolor[rgb]{0,0,0}{5}\\
\hline
\textcolor[rgb]{0,0,0}{\multirow{1}{*}{I-adder}} 
& number & \textcolor[rgb]{0,0,0}{12$\times$256}& 36.8 &\textcolor[rgb]{0,0,0}{40}\\
\hline
\multirow{1}{*}{\textcolor[rgb]{0,0,0}{ReLU}}  
& number     & 2& \textcolor[rgb]{0,0,0}{205} &300\\ 
\hline
\multirow{1}{*}{MaxPool} 
& number & 1& 330 &240\\
\hline
Input buffer & \multirow{1}{*}{size/\textcolor[rgb]{0,0,0}{number}} & \multirow{1}{*}{2KB/\textcolor[rgb]{0,0,0}{1}} & \multirow{1}{*}{12736} & \multirow{1}{*}{50}\\
\hline
Output buffer & \multirow{1}{*}{size/\textcolor[rgb]{0,0,0}{number}} & \multirow{1}{*}{2KB/\textcolor[rgb]{0,0,0}{1}} & \multirow{1}{*}{31039} & \multirow{1}{*}{50}\\
\hline
\textbf{Total} & \multicolumn{3}{c|}{} & \textbf{\textcolor[rgb]{0,0,0}{0.86~mm$^2$}}\\
\hline
\hline
\multicolumn{5}{|c|}{\textbf{\systemname~chip (\textcolor[rgb]{0,0,0}{40} MHz)}}\\
\hline
sub-Chip &  number     & 106\textcolor[rgb]{0,0,0}{$^a$}& ~ &\textcolor[rgb]{0,0,0}{0.86}~mm$^2$\\
\hline
\textbf{Total} & \multicolumn{3}{c|}{} & \textbf{\textcolor[rgb]{0,0,0}{91\textcolor[rgb]{0,0,0}{$^a$}}~mm$^2$}\\
\hline
\hline
\multicolumn{5}{|c|}{\textbf{Inter chips}}\\
\hline
\multirow{2}{*}{Hyper link} &  links/freq     & 1/1.6GHz& \multirow{2}{*}{1620} &\multirow{2}{*}{5.7~mm$^2$}\\
        & link bw         & 6.4 GB/s &~&  \\
\hline
\end{tabular}
\vspace{1pt}
\\\footnotesize{\scriptsize\textcolor[rgb]{0,0,0}{$^a$ Scaling TIMELY to an area of 0.86$\chi$ mm$^2$ by adjusting the number of sub-Chips (i.e., $\chi$) based on applications.}}
\vspace{-10pt}
\label{TIMEhardware}
\end{table}

\textbf{Methodology.}
We first compare \systemname~with \ul{4 state-of-the-art R$^2$PIM accelerators} (PRIME~\cite{PRIME}, ISAAC~\cite{ISAAC}, PipeLayer \cite{PipeLayer}, and AtomLayer~\cite{8465832}) in terms of peak energy efficiency and computational density. For this set of experiments, the performance data of the baselines are \textcolor[rgb]{0,0,0}{the ones reported} in their corresponding papers. Second, \textcolor[rgb]{0,0,0}{as for} the evaluation \textcolor[rgb]{0,0,0}{regarding} various benchmarks, we consider only PRIME~\cite{PRIME} and ISAAC~\cite{ISAAC} because (1) there is lack of design detail information to obtain results for PipeLayer \cite{PipeLayer} and AtomLayer~\cite{8465832}, and (2) more importantly, such comparison is sufficient given that PRIME~\cite{PRIME} is the most competitive baseline in terms of energy efficiency (see Fig. \ref{bottlenecks_state_of_the_arts} (c)). For this set of evaluations, \textcolor[rgb]{0,0,0}{we build an in-house simulator to evaluate the energy and throughput of PRIME, ISAAC, and \systemname. Before using our simulator, we validate it against PRIME's simulator~\cite{PRIME} and ISAAC's analytical calculations~\cite{ISAAC}. \textcolor[rgb]{0,0,0}{We set up our simulator to mimic PRIME and ISAAC and compare the results of our simulator with their original results}. The resulting errors of energy and throughput evaluation are 8\% and zero, respectively, which are acceptable by \systemname's one order of magnitude improvement on energy efficiency (see Section~\ref{final_result}).   
Due to the lack of ISAAC's mapping information, we only validate our simulator against PRIME's simulator to get the energy error by adopting PRIME's component parameters and weight mapping strategy~\cite{PRIME} in our simulator. Since PRIME does not support inter-layer pipeline, we only validate our simulator against ISAAC's analytical calculations to get the throughput error by using ISAAC's component parameters and balanced inter-layer pipeline~\cite{PRIME} in our simulator. The inter-layer pipeline corresponds to \systemname's inter-sub-Chip pipeline.}

\begin{figure*}[!t]
	\centering
	\vspace{-10pt}
	\includegraphics[width=172mm]{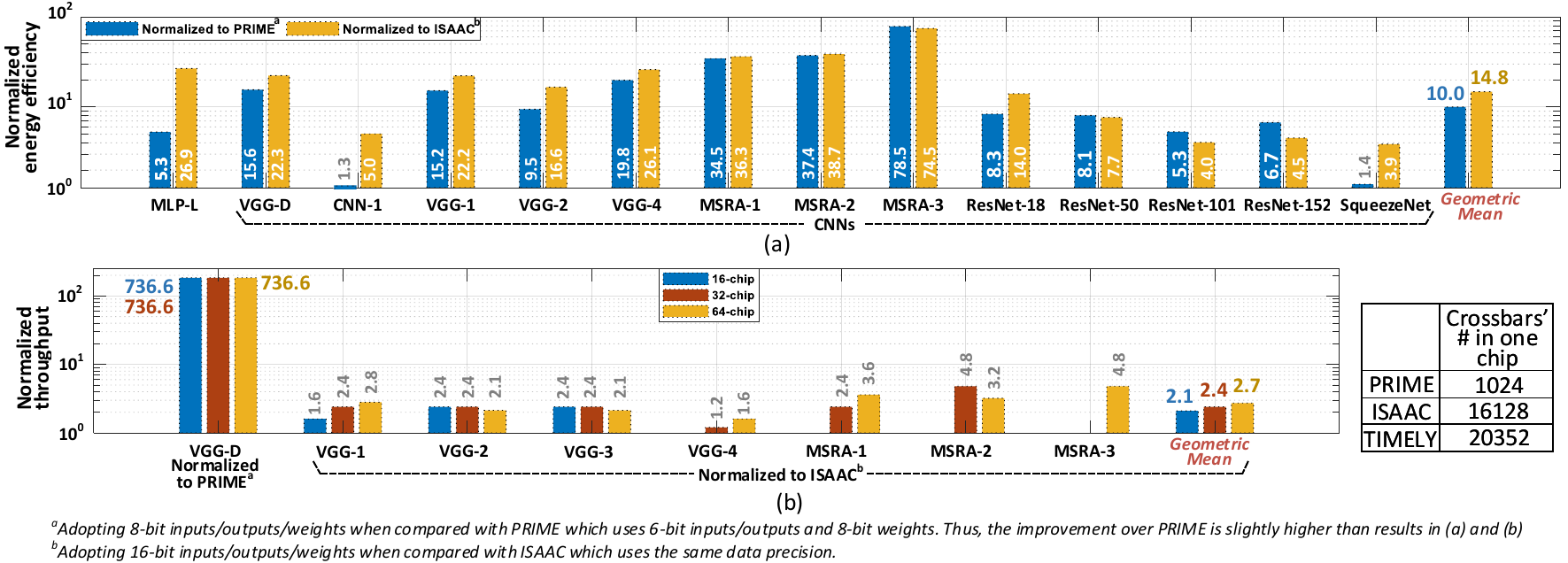}
	\vspace{-10pt}
	\caption{{(a) The normalized energy efficiency and (b) throughput of \systemname~over PRIME and ISAAC, respectively, considering various \textcolor[rgb]{0,0,0}{CNNs and DNNs}.}}
	\vspace{-13pt}
	\label{Performance on NN models}
\end{figure*}

\textbf{Benchmarks.} 
We evaluate \systemname using a total of \ul{15 benchmarks}. Table~\ref{benchmarkstable} shows these benchmarks and the reasons for adopting them.


\begin{table}[t!]
\scriptsize
\centering
\vspace{0pt}
\caption{\textcolor[rgb]{0,0,0}{Adopted benchmarks and datasets.} 
}
\vspace{-5pt}
\begin{tabular}{|c|c|}
\hline
\textbf{Benchmarks} & \textbf{Why consider these \textcolor[rgb]{0,0,0}{CNN/DNN} models}\\
\hline
\hline
\multirow{2}{*}{VGG-D$^a$, CNN-1$^b$, MLP-L$^b$} &  For a fair comparison with PRIME \\
~& (i.e. benchmarks in ~\cite{PRIME})\\
\hline
VGG-1/-2/-3/-4$^a$ &  For a fair comparison with ISAAC \\
 MSRA-1/-2/3$^a$ & (i.e. benchmarks in ~\cite{ISAAC})\\
\hline
ResNet-18/-50/-101/-152$^a$ & To show \systemname's performance \\
SqueezeNet$^a$ & in diverse and more recent CNNs \\
\hline
\end{tabular}
\\\footnotesize{\scriptsize{$^a$ ImageNet ILSVRC dataset~\cite{deng2009imagenet};}}
\footnotesize{\scriptsize{$^b$ MNIST dataset~\cite{MNIST}}}
\vspace{-15pt}
\label{benchmarkstable}
\end{table}

\subsection{Evaluation Results}
\label{final_result}
We evaluate \systemname's peak energy efficiency and computational density against those reported in \cite{PRIME},~\cite{ISAAC}, \textcolor[rgb]{0,0,0}{~\cite{PipeLayer}, and \cite{8465832}}. Next, we perform an  evaluation of \systemname's energy efficiency and throughput on various \textcolor[rgb]{0,0,0}{CNN and DNN} models. 

\begin{table}[h!]
\vspace{-5pt}
\centering
\caption{Peak performance comparison.}
\vspace{-5pt}
\scriptsize
\begin{tabular}{|p{1.55cm}<{\centering}|p{1cm}<{\centering}|p{1cm}<{\centering}|p{1.9cm}<{\centering}|p{1cm}<{\centering}|}
\hline
~& \textbf{Energy}\ & \textbf{Improve-} & \textbf{Computational} & \textbf{Improve-} \\
~& \textbf{efficiency}\ & \textbf{ment of} & \textbf{density} & \textbf{ment of} \\
~& $(TOPs/W)$\ & \textbf{\systemname}& $(TOPs/(s\times mm^2))$ &\textbf{\systemname} \\
\hline
\hline
PRIME$^a$~\cite{PRIME}& 2.10 & $+10.0\times$ & 1.23 & $+31.2\times$\\
\hline
ISAAC$^b$~\cite{ISAAC}& 0.38 & $+18.2\times$ & 0.48 & $+20.0\times$\\
\hline
{PipeLayer$^b$~\cite{PipeLayer}} & 0.14 & $+49.3\times$ & 1.49 & $+6.4\times$\\
\hline
{AtomLayer$^b$~\cite{8465832}} & 0.68 & $+10.1\times$ & 0.48 & $+20.0\times$\\
\hline
\textbf{\systemname$^a$} & \textcolor[rgb]{0,0,0}{\textbf{21.00}} & \small{n/a} & \textbf{38.33} & \small{n/a}\\
\hline
\textbf{\systemname$^b$} & \textcolor[rgb]{0,0,0}{\textbf{6.90}} & \small{n/a} & \textbf{9.58} & \small{n/a}\\
\hline
\end{tabular}
\\\footnotesize{\scriptsize{$^a$ one operation: 8-bit MAC;}}
\footnotesize{\scriptsize{$^b$ one operation: 16-bit MAC}}
\vspace{-20pt}
\label{Performance comparison_overview}
\end{table}
\vspace{-0pt}

\textbf{Overall Peak Performance.}
\textcolor[rgb]{0,0,0}{Compared} with representative R$^2$PIM accelerators (see Table~\ref{Performance comparison_overview}), \systemname~can improve  energy efficiency by over 10$\times$ (over PRIME~\cite{PRIME}) and the computational density by over 6.4$\times$ (over PipeLayer~\cite{PipeLayer}). \textcolor[rgb]{0,0,0}{In particular}, \systemname~improves energy efficiency by 10$\times$ to 49.3$\times$ and computational density by 6.4$\times$ to 31.2$\times$. These large improvements result from \systemname's innovative features of \textcolor[rgb]{0,0,0}{ALB}, TDI, O$^2$IR \textcolor[rgb]{0,0,0}{
and intra-sub-Chip pipelines}, which can aggressively reduce energy cost of the dominant data movements and increase the number of operations given the same \textcolor[rgb]{0,0,0}{time and} area. In Table~\ref{Performance comparison_overview}, we \textcolor[rgb]{0,0,0}{ensure} that \systemname's precision is the same as that of the baselines for a fair comparison. Specifically, we consider a 8-bit \systemname~design  when comparing with PRIME and a 16-bit \systemname~design when comparing to ISAAC, PipeLayer, and AtomLayer.

\begin{figure*}[!t]
	\centering
	\vspace{-10pt}
	\includegraphics[width=172mm]{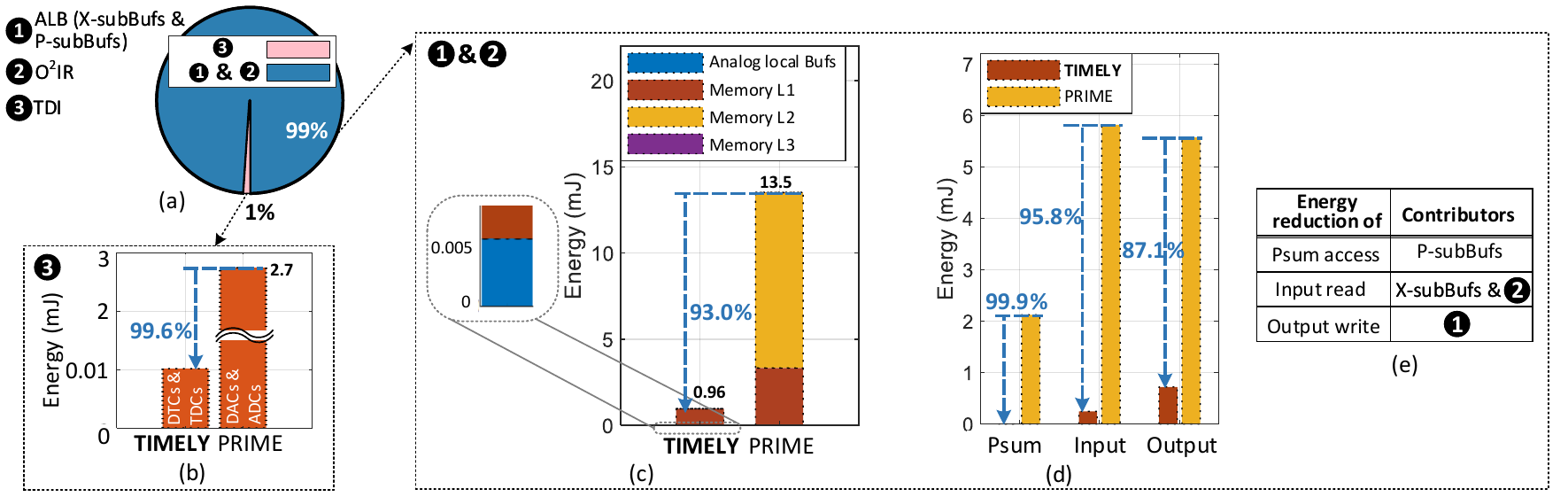}
	\vspace{-10pt}
	\caption{{The effectiveness of \systemname's innovations: (a) a breakdown of energy savings (over PRIME on VGG-D) achieved by different features \textcolor[rgb]{0,0,0}{-- i.e.} ({\color{black}{\ding{182}}} and {\color{black}{\ding{183}}}) vs. {\color{black}{\ding{184}}} of \systemname; (b) comparing \textcolor[rgb]{0,0,0}{the} energy cost\textcolor[rgb]{0,0,0}{s} of the interfacing circuits in \systemname~and PRIME; energy breakdown with regard to both (c) memory type\textcolor[rgb]{0,0,0}{s} and (d) data type\textcolor[rgb]{0,0,0}{s} in both \systemname~and PRIME; and (e) the contributing factors for the energy savings per data type in \systemname (see (d)).
\vspace{0pt}
}}
	\label{TIME_PRIME_interface}
	\vspace{-15pt}
\end{figure*}

\textbf{Energy Efficiency on Various \textcolor[rgb]{0,0,0}{CNN and DNN} models.} 
\textcolor[rgb]{0,0,0}{We} evaluate \systemname~on various models (\ul{1 MLP and 13 CNNs}) to validate that its superior performance is generalizable to different computational and data movement patterns. Fig. \ref{Performance on NN models} (a) shows the normalized energy efficiency of \systemname~over PRIME and ISAAC. We can see that \systemname~outperforms both PRIME and ISAAC on all \textcolor[rgb]{0,0,0}{CNN and DNN} models. Specifically, \systemname~is \ul{on average 10$\times$ and 14.8$\times$ more energy efficient than PRIME and ISAAC}, respectively (see the Geometric Mean in the rightmost part of Fig. \ref{Performance on NN models} (a)). This set of \textcolor[rgb]{0,0,0}{experimental} results demonstrates that \systemname's superior energy efficiency is independent \textcolor[rgb]{0,0,0}{of CNNs and DNNs -- i.e.} computational and data movement patterns. In addition, as shown in Fig. \ref{Performance on NN models} (a), the energy efficiency improvement of \systemname~\textcolor[rgb]{0,0,0}{decreases} in small or compact \textcolor[rgb]{0,0,0}{CNNs}, such as CNN-1 \cite{PRIME} and SqueezeNet \cite{2016squeezenet}. This is because their energy costs of data movements are relatively small. These models can be mapped into one ReRAM bank of PRIME or one ReRAM tile of ISAAC, and thus do not require high cost memory accesses and limit the energy savings achieved by \systemname.

\textbf{Throughput on Various \textcolor[rgb]{0,0,0}{CNNs}.} 
Fig. \ref{Performance on NN models} (b) shows \systemname's normalized throughput over PRIME and ISAAC on various \textcolor[rgb]{0,0,0}{CNNs} (\ul{a total of 8 CNNs}) considering three chip configurations (16, 32, and 64 chips). As the throughput is a function of the weight duplication ratio, we only consider CNNs for which PRIME or ISAAC provides corresponding weight duplication ratios. 
\ul{Compared to PRIME}, \systemname~enhances the throughput by 736.6$\times$ for the 16-chip, 32-chip, and 64-chip configurations on VGG-D. \textcolor[rgb]{0,0,0}{\systemname's advantageous throughput results from its intra-sub-Chip pipeline, which enables to minimize the latency between two pipelined outputs. In addition, PRIME can work in both the memory mode and computation mode (i.e. accelerating CNN), limiting the number of crossbars for CNN computations (and thus its throughput) on a chip which is over \textcolor[rgb]{0,0,0}{20$\times$} smaller than that of \systemname (i.e. 1024/20352, see the right corner of Fig. ~\ref{Performance on NN models} (b)).}
\ul{Compared to ISAAC on 7 CNNs}, TIMELY\textcolor[rgb]{0,0,0}{, on average,} enhances the throughput by 2.1$\times$, 2.4$\times$, and 2.7$\times$ for the 16-chip, 32-chip, and 64-chip configurations, respectively. In Fig. \ref{Performance on NN models} (b), we consider only 64-chip or (32-chip and 64-chip) for large \textcolor[rgb]{0,0,0}{CNNs}, such as MSRA-1/-2/-3, to ensure that all the models can be mapped into one \systemname~or ISAAC accelerator.
\systemname's enhanced throughput is because ISAAC adopts serial operations and requires 22 \textcolor[rgb]{0,0,0}{pipeline-}cycles (each being 100 ns) to finish one 16-bit MAC operation, for which \systemname~employs intra-sub-Chip pipelines and needs two \textcolor[rgb]{0,0,0}{pipeline-}cycles (each being 200 ns). 

\textbf{Accuracy.} 
We observe $\leq$ 0.1\%  inference accuracy loss under various CNN and DNN models in system-level simulations including circuit-level errors extracted from Cadence simulation. \textcolor[rgb]{0,0,0}{The simulation methodology is adapted from prior work~\cite{kang2015energy, kang2014energy}}. Specifically, \textcolor[rgb]{0,0,0}{ we first obtain noise and PVT variations (by Monte-Carlo simulations in Cadence) of  X-subBuf, P-subBuf, I-adder, DTC, and TDC.} The errors follow Gaussian noise distribution. We then add equivalent noise during training and use the trained weights for inference. Note that prior work has proved that adding Gaussian noise to training can reach negligible accuracy loss~\cite{polino2018model,morcos2018importance,santurkar2018does}. \textcolor[rgb]{0,0,0}{To achieve $\leq$ 0.1\% accuracy loss, we set the number of cascaded X-subBufs to 12. The accumulated error of the cascaded X-subBufs is $\sqrt{12}\varepsilon$~\cite{Accumulated_noise}, where $\varepsilon$ is the potential error of one X-subBuf. $\sqrt{12}\varepsilon$ is less than 20$\times$2$^8$ ps, which can be tolerated by the design margin of 40$\times$2$^8$ ps and thus do not cause a loss of inference accuracy.}

\vspace{-5pt}
\subsection{Effectiveness of \systemname's Innovations}
We first validate  the effectiveness of \systemname's innovations on energy saving and area reduction, and then \textcolor[rgb]{0,0,0}{demonstrate that}  \systemname's innovative principles can be generalized to state-of-the-art R$^2$PIM accelerators to further improve their energy efficiency.

\textbf{Effectiveness of \systemname's {I}nnovations \textcolor[rgb]{0,0,0}{on Energy Savings}.}
We here present an energy breakdown analysis to demonstrate how \systemname~reduces the energy consumption on VGG-D as compared with PRIME, which is the most competitive R$^2$PIM accelerator in terms of energy efficiency. In Fig. \ref{Performance on NN models} (a), we can see that \systemname~improves the energy efficiency by 15.6 $\times$ as compared to PRIME. 

\ul{Overview.} We first show the breakdown of energy savings achieved by different features of \systemname. \systemname's \textcolor[rgb]{0,0,0}{ALB} and O$^2$IR contribute to up to 99\% of the energy savings, and its \textcolor[rgb]{0,0,0}{TDI} 
leads to the remaining 1\% (see Fig. ~\ref{TIME_PRIME_interface} (a)).   

\ul{Effectiveness of \systemname's ALB and O$^2$IR.}
We compare \systemname's energy breakdown with regard to both memory types and data types with those of PRIME in Fig.~\ref{TIME_PRIME_interface} (c) and (d), respectively. 
In Fig.~\ref{TIME_PRIME_interface} (c), \systemname's \textcolor[rgb]{0,0,0}{ALB} and O$^2$IR together reduce the energy consumption of memory accesses by 93\% when compared with PRIME. Specifically, the \textcolor[rgb]{0,0,0}{ALB} and O$^2$IR features enable \systemname~to fully exploit local buffers within its \textcolor[rgb]{0,0,0}{sub-Chips} for minimizing accesses to the L1 memory and removing the need to access \textcolor[rgb]{0,0,0}{an} L2 memory.

In Fig.~\ref{TIME_PRIME_interface} (d), \systemname~reduces the energy consumption associated with the data movement of Psums, inputs and outputs by 99.9\%, 95.8\%, and 87.1\%, respectively\textcolor[rgb]{0,0,0}{. The} contributing factors are summarized in Fig.~\ref{TIME_PRIME_interface} (e). Specifically, (1) \systemname~can handle \textcolor[rgb]{0,0,0}{most of} the Psums locally via the P-subBufs within the \textcolor[rgb]{0,0,0}{sub-Chips}, aggressively reducing the energy cost of data movements of Psums; (2) \systemname's O$^2$\textcolor[rgb]{0,0,0}{IR} feature ensures all the input data are fetched only once from the L1 memory while its \textcolor[rgb]{0,0,0}{ALB} feature (i.e. X-subBufs here) allows the fetched inputs to be stored and transferred via X-subBufs \textcolor[rgb]{0,0,0}{between} the crossbars; and (3) thanks to employed P-subBufs and X-subBufs, \systemname~removes the need \textcolor[rgb]{0,0,0}{for an} L2 memory, which has $146.7\times/6.9\times$ higher read/write energy than that of \textcolor[rgb]{0,0,0}{an} L1 memory, respectively, reducing the energy cost of writing outputs back to the memory (L1 memory in \systemname vs. L2 memory in PRIME). 
Furthermore, as \ul{another way to see the effectiveness of \systemname's O$^2$IR feature}, we summarize both PRIME's and \systemname's total number of input accesses to the L1 Memory in Table~\ref{input_access_e_number} (consider the first six CONV layers as examples). \systemname~requires about 88.9\% less L1 memory accesses.

\begin{table}[b!]
\vspace{-15pt}
\centering
\caption{The total number of L1 memory accesses for reading inputs in \systemname~and PRIME~\cite{PRIME} considering VGG-D.}
\vspace{-5pt}
\scriptsize
\begin{tabular}{|p{1.1cm}<{\centering}|p{0.75cm}<{\centering}|p{0.85cm}<{\centering}|p{0.75cm}<{\centering}|p{0.85cm}<{\centering}|p{0.75cm}<{\centering}|p{0.75cm}<{\centering}|}
\hline
~& CONV1\ & CONV2 & CONV3 & CONV4 & CONV5 & CONV6 \\
\hline
\hline
\textcolor[rgb]{0,0,0}{PRIME~\cite{PRIME}} & 1.35 M & 28.90 M & 7.23 M & 14.45 M & 3.61 M & 7.23 M\\
\hline
\textbf{\systemname} & 0.15 M & 3.21 M & 0.80 M & 1.61 M & 0.40 M & 0.80 M\\
\hline
Save by & 88.9\% & 88.9\% & 88.9\% & 88.9\% & 88.9\% & 88.9\%\\
\hline
\end{tabular}
\label{input_access_e_number}
\end{table}

\begin{figure}[!b]
	\centering
	\vspace{-5pt}
	\includegraphics[width=86mm]{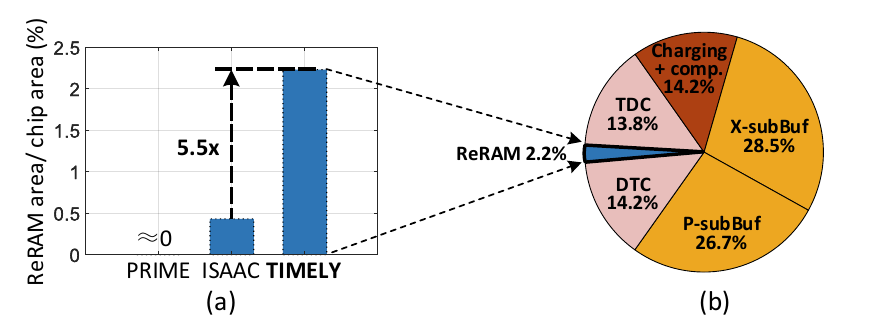}
	\vspace{-5pt}
	\caption{(a) The percentage of ReRAM crossbar area in the area of PRIME~\cite{PRIME}, ISAAC~\cite{ISAAC}, and TIMELY and (b) the area breakdown of \systemname}
	\label{area_breakdown}
\end{figure}

\ul{Effectiveness of \systemname's TDI.}
Although DTCs \textcolor[rgb]{0,0,0}{and} TDCs only contribute to 1\% of \systemname's energy savings over PRIME, the total energy of DTCs \textcolor[rgb]{0,0,0}{and} TDCs in \systemname~is 99.6\% less than that of ADCs \textcolor[rgb]{0,0,0}{and} DACs in PRIME (see Fig.~\ref{TIME_PRIME_interface} (b)). It is because (1) the unit energy of one DTC/TDC is about 30\%\textcolor[rgb]{0,0,0}{/}23\% of that of DAC/ADC; (2) the increased analog data locality due to ALBs largely reduces the need to activate DTCs and TDCs; and (3) \systemname's O$^2$IR feature aggressively reduces the required DTC conversions
thanks to its much reduced input accesses to the L1 memory (see Table~\ref{input_access_e_number}). 

\begin{figure}[!t]
 	\centering
 	\vspace{-0pt}
 	\includegraphics[width=86mm]{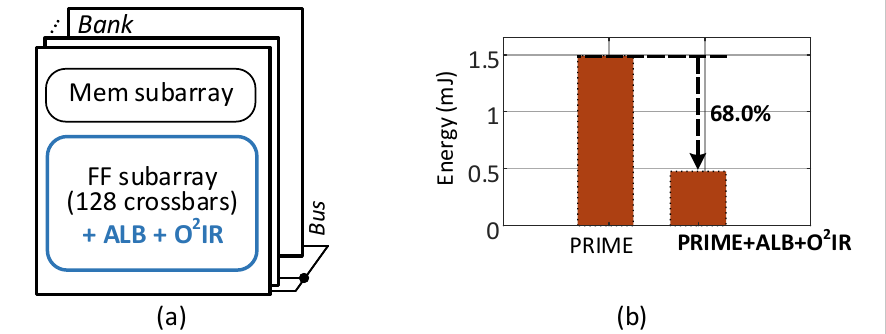}
 	\vspace{-10pt}
 	\caption{{Comparing the total energy cost of intra-bank data movements between PRIME and PRIME with \systemname's {ALB} and O$^2$IR being applied\textcolor[rgb]{0,0,0}{: (a) applying {ALB} and O$^2$IR to PRIME architecture and (b) the resulting energy reduction. }  
 	}}
 	\label{PRIME_analog_buffer}
 	\vspace{-15pt}
\end{figure}

\textbf{{E}ffectiveness of \systemname's {I}nnovations \textcolor[rgb]{0,0,0}{on Area Reduction.}}
\textcolor[rgb]{0,0,0}{We analyze an area breakdown to present the effectiveness of \systemname's innovations on the area savings of peripheral circuits, which helps to improve the computational density (see Table~\ref{Performance comparison_overview}). In Fig.~\ref{area_breakdown} (a), the percentage of the ReRAM array area  in \systemname (i.e. 2.2\%) is 5.5$\times$ higher than that in ISAAC (i.e. 0.4\%)~\cite{ISAAC}. The percentage of the ReRAM array area in PRIME is small enough and thus ignored~\cite{PRIME}. The higher percentage of the ReRAM crossbar array area in \systemname benefits from area-efficient circuit implementations of \systemname's ALB, TDI and O$^2$IR. Specifically, in \systemname shown in Fig.~\ref{area_breakdown} (b), X-subBufs and P-subBufs occupy 55.2\% of the chip area; DTCs and TDCs occupy 28\% of the chip area; the area of CMOS logic introduced by O$^2$IR is neglectable.}

\vspace{0pt}
\textcolor[rgb]{0,0,0}{\textbf{Generalization of \systemname's Innovations.}}
\textcolor[rgb]{0,0,0}{\systemname's innovative features} are generalizable and can be applied to state-of-the-art R$^2$PIM accelerators for boosting their energy efficiency. 
\textcolor[rgb]{0,0,0}{To demonstrate, we apply ALB and O$^2$IR to PRIME based on the following considerations. ALB feature associated with O$^2$IR contributes the dominant energy savings (see Fig.~\ref{TIME_PRIME_interface} (a)). From the perspective of data accesses and interfaces, PRIME uses the same architecture shown in Fig~\ref{TIME-innovation} (a) as
ISAAC~\cite{ISAAC}/PipeLayer~\cite{PipeLayer}. To evaluate, we modify PRIME architecture as shown in Fig.~\ref{PRIME_analog_buffer} (a). 
%
We add X-subBufs and P-subBufs between 128 ReRAM crossbar arrays in FF subarray of each bank, and modify the weights mapping and input access dataflow based on O$^2$IR, while employing PRIME's original designs outside FF subarray. 
Thus, ALB and O$^2$IR only have an impact on the intra-bank energy. In this experiment, we adopt the same component parameters as those used in the PRIME's original design.}
Fig.~\ref{PRIME_analog_buffer} \textcolor[rgb]{0,0,0}{(b)} shows that applying \textcolor[rgb]{0,0,0}{ALB} and O$^2$IR principle to FF subarrays in PRIME reduces the intra-bank data movement energy by 68\%. 

\vspace{-15pt}
\textcolor[rgb]{0,0,0}{\subsection{Discussion}}
\textcolor[rgb]{0,0,0}{Area scaling of TIMELY (by adjusting the number of sub-Chips shown in Table~\ref{TIMEhardware}) does not affect throughput and slightly affects energy. This is because throughput is determined only by intra-sub-Chip pipeline (see Section~\ref{pipeline_arch}); adjusting the number of sub-Chip in one chip will only change inter-chip energy (i.e., the energy of memory L3  in Fig. ~\ref{TIME_PRIME_interface} (c)), which accounts for a negligible part of the total energy.}

\vspace{4pt}
\section{Related Work}
\textbf{Non-PIM \textcolor[rgb]{0,0,0}{CNN/}DNN \textcolor[rgb]{0,0,0}{A}ccelerators.} Although memory is only used for data storage in non-PIM accelerators, comput\textcolor[rgb]{0,0,0}{ing} units are being pushed closer to compact memories to reduce energy and area. For accelerators with off-chip DRAM, DRAM accesses consume two orders of magnitude more energy than on-chip memory accesses (e.g. 130$\times$ higher than a 32-KB cache at 45 nm~\cite{6757323}). As a result, DRAM consumes more than 95\% of the total energy in DianNao~\cite{DianNao,7284058}. To \textcolor[rgb]{0,0,0}{break through} the off-chip bottleneck, on-chip SRAM is widely used as the mainstream on-chip memory solution~\cite{Han_SISCA_16}. However,  SRAM's low density has been limiting its on-die capacity even with technology scaling. For example, EIE adopts 10-MB SRAM that takes 93.2\% of the total area~\cite{Han_SISCA_16}. To address the area issue, on-chip eDRAM is used in RANA~\cite{8416839} and DaDianNao~\cite{7011421}, as eDRAM can save about 74\% area \textcolor[rgb]{0,0,0}{while providing} 32 KB capacity in 65 nm~\cite{8416839,7092634}. However, the refresh energy in eDRAM can be dominant (e.g. \textcolor[rgb]{0,0,0}{about 10$\times$ as high as} the data access' energy~\cite{8416839}). In terms of FPGA-based designs, the performance is also limited by the memory accesses~\cite{6861585,7092634,8310398,Fpga3} with limited flexibility of choosing memory technologies. Different from these non-PIM accelerators, \systemname~improves energy efficiency by computing in memory and enhances computational density through \textcolor[rgb]{0,0,0}{adopting} high-density ReRAM.

\vspace{0pt}
\textbf{PIM \textcolor[rgb]{0,0,0}{CNN/}DNN \textcolor[rgb]{0,0,0}{A}ccelerators.}
While PIM accelerators integrate comput\textcolor[rgb]{0,0,0}{ing} units in memory to save the energy \textcolor[rgb]{0,0,0}{of accessing weights}, the achievable energy efficiency and computational density \textcolor[rgb]{0,0,0}{remain} limited. 
\textcolor[rgb]{0,0,0}{The limited energy efficiency is induced by} the energy cost of input and Psum movements and the overhead of interfacing circuits. PRIME~\cite{PRIME} takes 83\% of the total energy to access inputs and Psums, and  ISAAC~\cite{ISAAC} consumes 61\% of the total energy to operate DACs/ADCs.
\textcolor[rgb]{0,0,0}{The limited computational density is related to memory technologies.} Processing in SRAM, for example, faces \textcolor[rgb]{0,0,0}{this} limitation. \textcolor[rgb]{0,0,0}{The reasons include not only one SRAM bit-cell typically stores only 1-bit weight~\cite{8310397,8310398,8310401,Sandwich-PIsram-ISSCC19,8T-PIsram-Mengfan-ISSCC19} but also SRAM's bit-cell structure -- e.g. 6T~\cite{8310398,8310401}/8T~\cite{Sandwich-PIsram-ISSCC19,8T-PIsram-Mengfan-ISSCC19}/10T~\cite{8310397} structure -- decreases density.}
\textcolor[rgb]{0,0,0}{Proposed} \systemname~\textcolor[rgb]{0,0,0}{adopts high-density ReRAM, and} addresses two key \textcolor[rgb]{0,0,0}{energy} challenges \textcolor[rgb]{0,0,0}{with} techniques including \textcolor[rgb]{0,0,0}{ALBs, TDIs, and O$^2$IR.} {\systemname~achieves up to 18.2$\times$ improvement (over ISAAC) in energy efficiency, 31.2$\times$ improvement (over PRIME) in computational density, and 736.6$\times$ in throughput (over PRIME).} \textcolor[rgb]{0,0,0}{Similar to the effect of ALBs used in \systemname, a recent R$^2$PIM accelerator~\cite{CASCADE} also increases the amount of data in the analog domain for energy optimization. However, it only optimizes the computation energy (including the energy of interfacing circuits).}
\vspace{-5pt}
\section{Conclusions}
\vspace{-4pt}

In this paper, we analyze existing designs of R$^2$PIM accelerators and
identify three opportunities to greatly enhance their energy efficiency: analog data locality, time-domain interfacing, and input access reduction. \textcolor[rgb]{0,0,0}{These three opportunities inspire three key features of \systemname:} 
(1) ALBs, (2) interfacing with TDCs/DTCs, and (3) an O$^2$IR mapping method.
\systemname outperforms state-of-the-art in both energy efficiency and computational density while maintaining a better throughput.

\vspace{-10pt}
\section*{Acknowledgment}
\vspace{-4pt}
The authors would like to thank Dr. Rajeev Balasubramonian and Dr. Anirban Nag for their discussions on ISAAC~\cite{ISAAC}.


\bibliographystyle{IEEEtranS}


\end{document}